\PassOptionsToPackage{table}{xcolor}
\PassOptionsToPackage{capitalize,noabbrev}{cleveref}
\documentclass{bytedance_seed_arxiv}

\usepackage[utf8]{inputenc}   
\usepackage[T1]{fontenc}      
\usepackage{hyperref}         
\usepackage{xcolor}         
\definecolor{citecolor}{HTML}{2980b9}
\definecolor{linkcolor}{HTML}{c0392b}
\usepackage{url}              
\usepackage{xurl}             
\usepackage{microtype}        

\usepackage{amsmath}
\usepackage{amssymb}
\usepackage{amsfonts}
\usepackage{mathtools}
\usepackage{amsthm}
\usepackage{bm}
\usepackage{nicefrac}

\usepackage{booktabs}         
\usepackage{graphicx}
\usepackage{subcaption}
\usepackage{multirow}
\usepackage{makecell}
\usepackage{nicematrix}
\usepackage{caption}
\usepackage[table]{xcolor}
\usepackage{wrapfig}
\usepackage{enumitem}
\usepackage[most]{tcolorbox}
\usepackage{pifont}

\usepackage{algorithm}
\usepackage{algorithmic}

\usepackage[capitalize,noabbrev]{cleveref}
\usepackage{xspace}  

\theoremstyle{plain}

\theoremstyle{definition}

\theoremstyle{remark}

\newcommand{\method}{SpectraReward\xspace}

\newlength\savewidth
\newcommand{\tablestyle}[2]{\setlength{\tabcolsep}{#1}\renewcommand{\arraystretch}{#2}\centering\footnotesize}

\definecolor{myblue}{RGB}{232, 242, 255}
\definecolor{lightred}{RGB}{255, 182, 193}
\definecolor{lightgreen}{RGB}{144, 238, 144}
\definecolor{lightblue}{RGB}{173, 216, 230}
\definecolor{lightergray}{gray}{0.90}

\title{Read It Back: Pretrained MLLMs Are Zero-Shot Reward Models for Text-to-Image Generation}
\author[1]{Runhui Huang}
\author[3]{Qihui Zhang}
\author[1]{Zhe Liu}
\author[2]{Yu Gao}
\author[2]{Jie Wu}
\author[1]{Hengshuang Zhao}
\affiliation[1]{The University of Hong Kong}
\affiliation[2]{ByteDance Seed}
\affiliation[3]{Peking University}
\date{July 13, 2026}
\checkdata[Project Page]{\url{https://huangrh99.github.io/projects/SpectraReward/}}
\fancypagestyle{firststyle}{%
  \fancyhf{}%
  \fancyhead[L]{\vskip 1mm\raisebox{-5mm}[0pt][0pt]{\includegraphics[height=15mm]{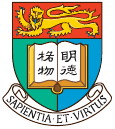}}}%
  \fancyhead[R]{\vskip 4mm\includegraphics[width=53.3mm,height=6mm]{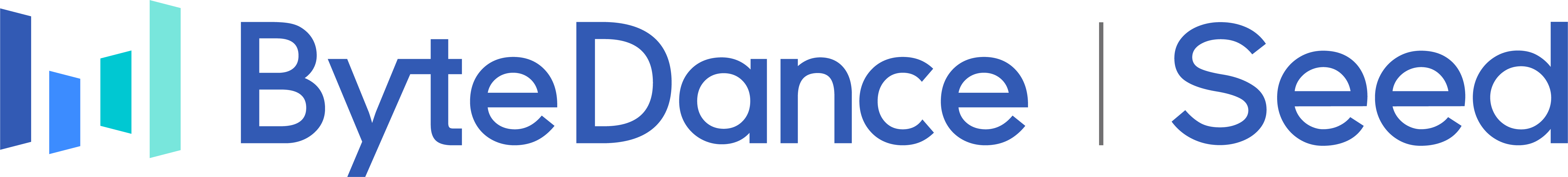}}%
}
\renewcommand{\seedanceteambadge}{}

\let\seedabstract\abstract
\RenewDocumentEnvironment{abstract}{+b}{\seedabstract{#1}}{}
\begin{abstract}
In this paper, we propose \textbf{\method,} a training-free reward function that turns pretrained MLLMs into off-the-shelf reward models
for image-generation reinforcement learning.
Instead of asking the MLLM to judge a generated image or answer decomposed verification questions,
\method measures how well the original prompt can be recovered from the generated image through a single image-conditioned, teacher-forced forward pass.
We use the average image-conditioned prompt log-likelihood as the reward, directly reusing the MLLM's pretrained image-text alignment ability without
preference labels, reward-model fine-tuning.
We further introduce \textbf{Self-\method,} a special case for unified multimodal models where the policy's own understanding branch serves as the reward model for its
generation branch, forming a closed-loop self-improving framework without external reward models or external knowledge.
Extensive experiments validate \method through a broad image-generation RL study covering two diffusion models, three RL algorithms,
nine reward MLLM backbones from four MLLM families spanning 4B to 235B parameters, and five out-of-distribution text-to-image benchmarks.
Results show that both \method and Self-\method significantly and consistently improve generation performance and outperform prior MLLM-derived reward training methods.
Further analysis reveals that larger reward MLLMs are not always better, while Self-\method can match or surpass much larger external reward models,
suggesting that reward-policy alignment is a key factor for effective image-generation RL.
\end{abstract}

\begin{document}
\maketitle

\renewcommand{\subsectionautorefname}{Section}
\renewcommand{\sectionautorefname}{Section}
\section{Introduction}
\label{sec:intro}

Image generation has advanced rapidly in recent years, evolving from
specialized text-to-image models~\cite{sd3,flux} to unified multimodal
models (UMMs)~\cite{bagel,chameleon,januspro,emu3,emu3.5,illume+,metaquery}
that integrate visual understanding and generation within a single
architecture. Reinforcement learning has emerged in parallel as an
effective post-training stage~\cite{flowgrpo,dancegrpo,awm,diffusionnft},
consistently lifting compositional fidelity and instruction-following. The success of an RL recipe, however, rests on two
complementary pillars. 
The optimization algorithm governs the stability of long-horizon training,
while the reward model determines the ceiling that the trained policy can
ultimately approach. 
However, designing a practical reward model that remains both efficient and reliable is still challenging.

Recent studies have made substantial progress in reward modeling for image generation.
One line of work builds reward models from large-scale
human preference annotations, using these data to align visual-text
representations or vision-language models with human judgments~\cite{pickscore,imagereward,hpsv2,rewarddance,unifiedreward}. While effective, these methods depend on expensive annotation, difficult data collection, and costly iteration. Another line of work bypasses preference training by reusing pretrained MLLMs as zero-shot reward sources~\cite{viescore,li2025uniworld,alphagrpo}. Direct scalar or logit-based feedback is training-free, but can be sensitive to judge calibration and scoring noise~\cite{viescore,li2025uniworld}. Question-decomposition pipelines improve reliability by verifying atomic prompt requirements one by one, but introduce substantial engineering complexity~\cite{alphagrpo}. This leaves open the design of an efficient, preference-label-free, and off-the-shelf reward model for open-sourced image generators.

\begin{figure*}[t]
  \centering
  \includegraphics[width=\linewidth]{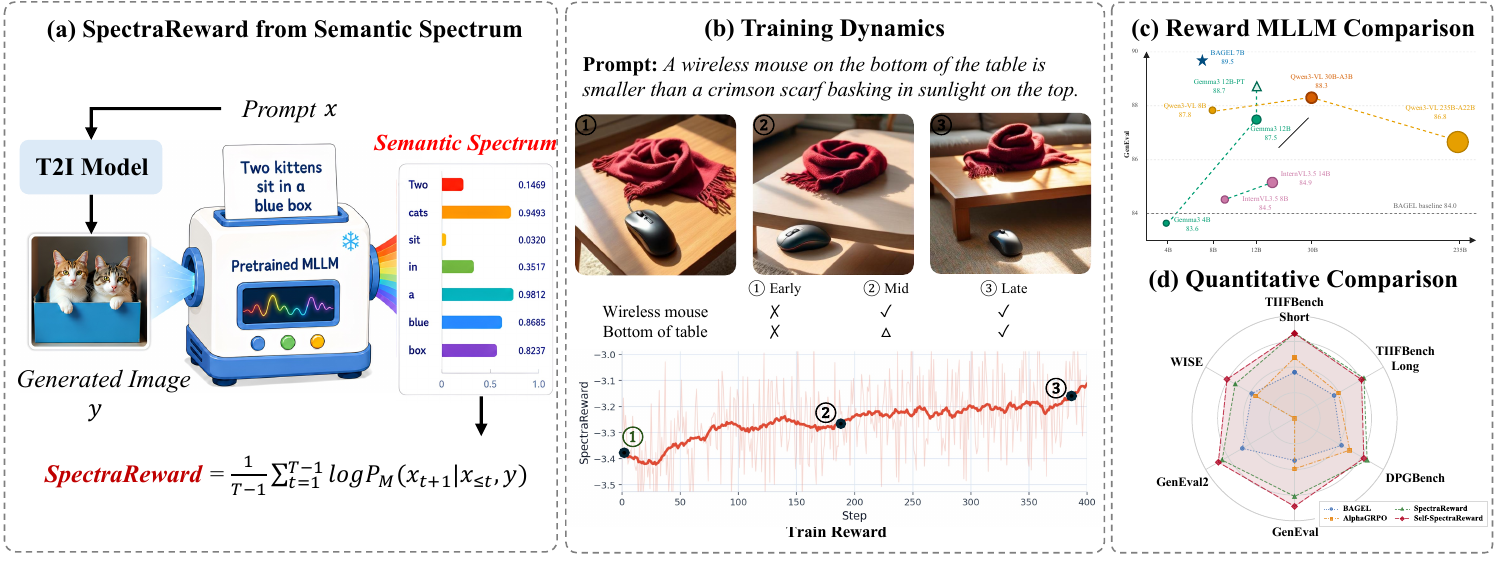}
    \vspace{-10pt}
  \caption{\textbf{Overview of \method.}
  (a) Pretrained MLLMs naturally induce a semantic spectrum that measures how well a generated image aligns with the prompt.
  \method aggregates this into a reward for T2I RL.
  (b) During RL training, \method steadily increases together with visible improvements in complex scene generation.
  (c) We study nine reward MLLM backbones from four MLLM families, with external reward MLLMs spanning three families and 4B to 235B parameters. 
  Scaling the reward MLLM backbones brings non-monotonic gains.
  Qwen3-VL-30B-A3B achieves the best performance among external MLLMs,
  while Self-\method, using BAGEL's own understanding branch as the reward model, outperforms all external MLLMs.
  (d) Both \method and Self-\method bring significant and consistent improvements across all five downstream benchmarks compared to the baselines.
  }
  \label{fig:teaser}
  \vspace{-10pt}
\end{figure*}

We propose \method, a training-free reward function that turns any
pretrained MLLM into an image-generation reward model by reusing its
image-text alignment knowledge, without preference labels or additional training. 
Given a generated image, \method conditions a frozen MLLM on the image and performs a single teacher-forced forward pass on the prompt, 
producing a token-level likelihood profile that reflects how well each textual requirement can be read from the image. 
We refer to this profile as the \emph{semantic spectrum} of the image-text pair, 
and aggregate it into a scalar reward using the mean prompt-token log-likelihood. 
In this way, \method uses the prompt-likelihood spectrum induced by the model's own image-conditioned language modeling objective as a deterministic, off-the-shelf reward signal, as shown in~\autoref{fig:teaser}~(a), instead of asking the MLLM to judge an image or answer decomposed questions.

We further specialize \method within unified multimodal models (UMMs),
yielding a special case we call \emph{Self-\method}. Since a UMM integrates image
understanding and image generation into a single framework, its 
understanding branch could directly provide the reward signals for its
generation branch.
In this way, the model improves its own generation
using its own understanding ability, without relying on any external
model or external knowledge, forming a closed-loop self-improving framework in which the policy evolves by
leveraging its own intrinsic capability. 
Beyond removing external dependencies, this design has a structural advantage. 
The understanding branch shares the same tokenizer, vision encoder, and pretraining
distribution as the generation branch, so the way it reads an image is naturally aligned with 
the distribution from which the generation branch samples images. 
We therefore hypothesize that this intrinsic alignment makes the policy's own understanding branch a
particularly suitable reward source for its own generation.

We validate \method through a broad image-generation RL study covering
two diffusion models, three RL training algorithms, nine reward MLLM backbones from four MLLM
families ranging from 4B to 235B parameters, and evaluate real generalization improvements on five out-of-distribution downstream text-to-image generation benchmarks.
On both SD3.5-M~\cite{sd3} and BAGEL~\cite{bagel}, \method brings significant improvements on all downstream benchmarks, showing the generalization ability of the proposed prompt-likelihood reward.
With BAGEL as the policy model, both \method and Self-\method consistently improve the BAGEL baseline by +10.0 on TIIF-Bench~\cite{tiif} and +4.3/5.5 on GenEval~\cite{geneval}, respectively.
\method also outperforms the strong RL baseline, AlphaGRPO~\cite{alphagrpo}, by 6.3 and 3.3 gains on TIIF-Bench and GenEval, respectively.
Moreover, Self-\method exhibits better improvements compared to \method with similar-scale MLLM backbones, and comparable or even better performance compared to 4x or 30x larger MLLMs,
e.g., +1.2 on GenEval, +2.1 on GenEval2~\cite{kamath2025geneval}, and +2 on WISE~\cite{wise} compared to \method with the best MLLM backbone.
These results support our hypothesis that distributional alignment between the reward model and the policy can be
as important as the scale of the reward model. The contributions of the paper are summarized as follows:
\begin{itemize}[leftmargin=*, itemsep=1pt, topsep=1pt]
  \item We propose \method, a training-free and off-the-shelf reward
  function that turns any pretrained MLLM into a reward model for text-to-image generation. 
  By calculating the image-conditioned prompt-token likelihood, 
  \method reuses pretrained image-text alignment without preference labels, 
  reward-model fine-tuning, scalar judging, or question decomposition.

  \item We introduce {Self-\method}, the special case in which the policy's own understanding branch in a unified multimodal model serves as the reward model. 
  This forms a closed-loop self-improving framework without external reward models or 
  external knowledge, where the model leverages its own understanding ability to improve generation performance.

  \item We conduct a broad image-generation RL study covering two
  generator backbones, three RL algorithms, nine reward MLLM backbones
  from four MLLM families spanning $4$B--$235$B parameters, and five
  downstream benchmarks. The results show the effectiveness of \method and Self-\method, and reveal that the benefits of scaling the reward model are not monotonic
  and that the well-aligned self-reward model can match or even outperform much larger
  external models.
\end{itemize}

\section{Related Work}
\label{sec:related_work}
\noindent \textbf{Reward Models for Image Generation.}
Reinforcement learning from human feedback (RLHF)~\cite{ouyang2022training} is a standard paradigm for aligning LLMs~\cite{gao2024honestllm,shao2024deepseekmath}, and has been adapted to T2I models~\cite{black2023training} via reward models trained on extensive human preferences or prompt–image alignment data~\cite{imagereward, visionreward, pickscore, imagereward, hpsv3}. 
While these reward models provide critical alignment signals, their effectiveness is fundamentally bottlenecked by an inherent bias against text-independent aesthetic details~\cite{ba2025enhancing}. 
To overcome these capacity limits, a second line of research upgrades the reward architecture to fine-tuned Multimodal Large Language Models (MLLMs), which successfully raise the performance ceiling~\cite{visionreward, rewarddance, gong2025onereward}. 
However, this requires large amounts of annotated data, substantial training overhead, and an inevitable shift in distribution away from the MLLM's pretraining knowledge. More recently, an emerging paradigm dispenses with both preference labels and reward-model fine-tuning. 
This approach leverages pretrained MLLMs as zero-shot judges, capable of assessing image quality and aesthetic reasoning directly~\cite{jiang2025multimodal, schoepp2025evolving}. These methods typically operate either by emitting a direct scalar score or by decomposing the generation prompt into verifiable atomic questions answered by the MLLM~\cite{alphagrpo}. 
Caption likelihood was used for image self-reranking in CogView~\cite{ding2021cogview}.
While concurrent work such as PromptEcho~\cite{liu2026promptecho} explores a cross-entropy function of this concept, its evaluation is restricted to domain-specific T2I backbones.

\noindent \textbf{Self-Rewarding in Unified Multimodal Models.}
Self-rewarding mechanisms, wherein a model leverages its own evaluations to supervise its training, have demonstrated significant efficacy in LLM alignment~\cite{yuan2024self, wu2025meta, sun2023salmon}. This paradigm naturally extends to Unified Multimodal Models (UMMs)~\cite{team2024chameleon, xie2024show, wu2024next, zhao2025unified}, which inherently integrate both understanding and generation capabilities. 
Alongside broader efforts to align multimodal models using self-generated preference data and iterative self-evolution~\cite{tan2025beyond, jiang2024modality}, 
recent studies have explored the specific intersection of self-rewarding and UMMs~\cite{jin2026latentum}: SRUM~\cite{jin2025srum} introduces an aggregation of global and local signals from the UMM's comprehension branch, 
while GvU~\cite{pan2026learning} utilizes the token-level logits of the understanding branch as dense reward signals. 
Although these studies align with the intuition that a policy's internal comprehension can effectively guide its own generation, 
they are confined to UMM-internal evaluations and fail to investigate how this reward compares against external reward models.

\section{Method}
\label{sec:method}

\begin{figure*}[t]
  \centering
  \includegraphics[width=\linewidth]{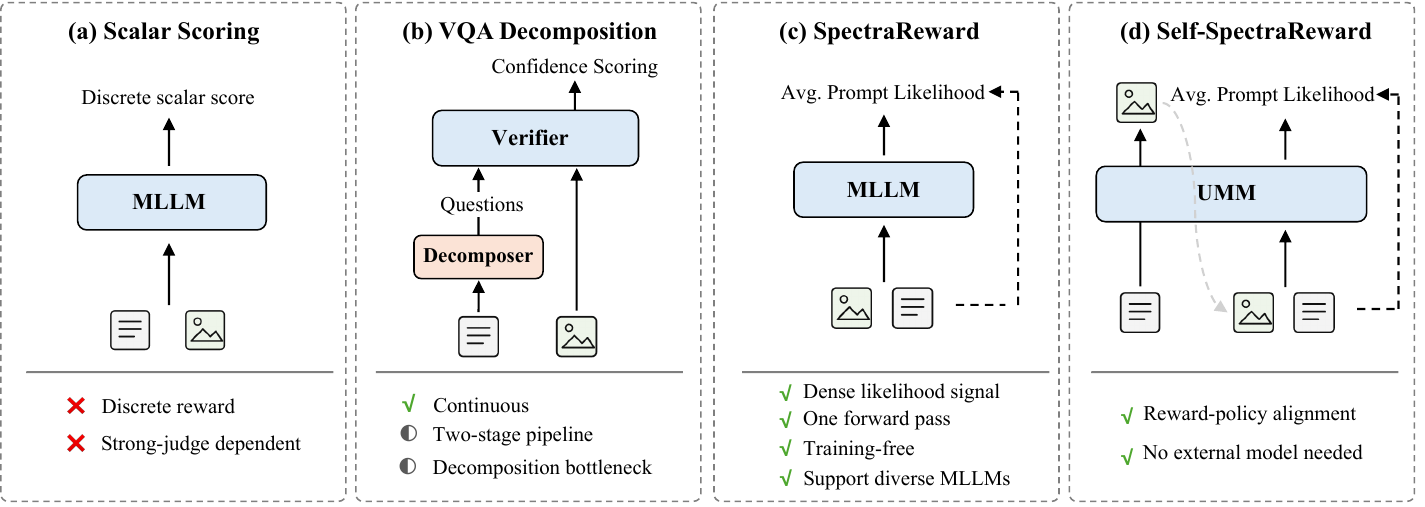}
  \caption{\textbf{Comparison of MLLM-based reward functions.}
  (a) Scalar scoring directly asks an MLLM to assign a discrete
  image-text alignment score, making the reward sensitive to judge
  calibration and scoring noise.
  (b) VQA decomposition converts the prompt into atomic questions and
  aggregates verifier confidence, but introduces a two-stage pipeline and
  depends on the quality of question decomposition.
  (c) \method computes the image-conditioned prompt likelihood through a single teacher-forced forward pass. 
  The resulting token-level likelihoods form a semantic spectrum, whose average is used as the scalar reward.
  (d) Self-\method instantiates the same prompt-likelihood reward within a unified multimodal model by using the policy's own understanding branch,
  removing the need for an external reward MLLM and improving reward-policy alignment.
  }
  \label{fig:method}
    \vspace{-10pt}
\end{figure*}
In this section, we first introduce the \method reward function and its token-level advantage variant in~\autoref{sec:reward_def}.
We then specialize it to unified multimodal models in~\autoref{sec:self_reward_def}, yielding Self-\method.
Finally,~\autoref{sec:analysis} analyzes the
discriminative properties of the reward signal.~\autoref{fig:method} compares \method with previous MLLM-based reward functions.

\subsection{\method}
\label{sec:reward_def}
Let $G_\theta$ be a text-to-image policy that samples an image $y \sim G_\theta(\cdot \mid x)$ 
from a prompt $x = (x_1, \ldots, x_T)$, and let $\mathcal{M}$ be the frozen pretrained MLLM.
Since pretrained MLLMs already learn strong image-text alignment from large-scale pretraining,
our goal is to extract a scalar reward $R_{\mathcal{M}}(x,y)$ from the pretrained MLLMs
that measures how well the generated image $y$ realizes the prompt $x$, 
while requiring no preference labels, reward-model fine-tuning, or other auxiliary pipeline.

The key idea of \method is to evaluate the generated image by asking a simple inverse question: 
how well the generated image $y$ can be translated back into the prompt $x$ from the perspective of $\mathcal{M}$.
To compute this, we feed the image $y$ as the visual condition
to $\mathcal{M}$, and then perform a single teacher-forced forward pass over the prompt tokens $x$.
The final reward function is as follows:
\begin{equation}
  \label{eq:method_reward}
  R_{\mathcal{M}}(x,y)
  =
  \frac{1}{T-1}
  \sum_{t=1}^{T-1}
  \log p_{\mathcal{M}}(x_{t+1} \mid x_{\le t}, y).
\end{equation}
This reward is the mean image-conditioned prompt log-likelihood under $\mathcal{M}$. 
The token-wise likelihoods form the \emph{semantic spectrum} of the image-text pair, 
describing how the visual evidence in $y$ supports the semantics of $x$ across prompt tokens.
\method aggregates this semantic spectrum into a scalar reward, where a higher reward $R(x,y)$ indicates that 
the prompt is more predictable from the image and that the generated image better supports the semantic content of the prompt.

The key advantage is that this reward function directly invokes the most fundamental capability of pretrained
MLLMs, namely understanding the image and associating it with the corresponding language description.
Thus, rather than training a separate reward model or prompting the MLLM to produce an abstract judgment score, 
\method reuses the fundamental visual-language grounding ability already learned during MLLM pretraining.

\noindent \textbf{Token-level advantage variant.}
A concern with long prompts is that averaging many token log-likelihoods may dilute the signal of a localized visual error.
We therefore consider an optional variant of \method that normalizes token-level scores before aggregation.
For a fixed prompt with $L$ scored tokens, let $\ell_{i,t}$ be the log-likelihood of token $t$ for rollout $i$, and let $\bar\ell_t$ be its mean across rollouts of that prompt.
The default first forms the sequence-level reward $R_i=L^{-1}\sum_t\ell_{i,t}$ and then computes a group-relative advantage.
The token-level variant instead computes
\begin{equation}
\label{eq:token_advantage}
A_i^{\mathrm{tok}}=\frac{1}{L}\sum_{t=1}^{L}
\frac{\ell_{i,t}-\bar\ell_t}{s_t+\epsilon},
\end{equation}
where $\epsilon>0$ ensures numerical stability.
We consider three standard-deviation scopes: global std over all scored tokens, group std over scored tokens within the same prompt group, and per-token std across rollouts at each position.
The first two share a standard deviation across token positions; the third uses a position-specific $s_t$.
All variants produce one image-level advantage for policy optimization.
With a shared $s_t=s$, this reduces to $A_i^{\mathrm{tok}}=(R_i-\bar R)/(s+\epsilon)$, where $\bar R$ is the mean reward within the prompt group.
Thus, centering cancels image-independent token contributions and dynamically emphasizes tokens that vary more across rollouts, which carry more ranking information.
The default shares this property, so a shared standard deviation only rescales the advantage.
Per-token normalization instead rescales each position to a comparable variance, which reduces this emphasis, and image-dependent variation can still obscure localized errors.
We evaluate these choices in~\autoref{sec:ablation_token}.

\subsection{Self-\method}
\label{sec:self_reward_def}
When $G_\theta$ is a unified multimodal model, it contains both a
generation branch $G_\theta^{\mathrm{gen}}$ and an understanding branch
$G_\theta^{\mathrm{und}}$. Self-\method instantiates \method by using
$G_\theta^{\mathrm{und}}$ as the reward MLLM for images sampled by $G_\theta^{\mathrm{gen}}$. 
In other words, the model scores its own generated images by measuring 
how likely the original prompt is under its own image-conditioned understanding branch.
This forms a closed-loop self-improving framework in which the model uses its own understanding ability to improve the generation quality.
Moreover, this self-rewarding design eliminates the need for external reward models,
avoids using additional computational resources to serve large-scale MLLMs, and simplifies the RL training pipeline.

\noindent \textbf{Reward-policy alignment.}
Self-\method couples two dual capabilities within the same unified model.
The generation branch maps text to images, while the reward function
uses the understanding branch to measure how well the generated image can be translated back to the original text. 
Since the two branches share the same tokenizer, vision encoder, and pretraining distribution, 
the reward signal is compatible with the policy's own generation knowledge. 
This does not require the understanding branch to be stronger than external MLLMs in general visual understanding. 
Instead, it provides a reward signal that is better calibrated to the policy's own generated-image distribution. 
Optimizing the generation branch with this self-reward therefore encourages the text-to-image policy to become more consistent
with its own image-to-text understanding ability. 
We further analyze the reward-policy alignment effect in~\autoref{sec:reward_backbone}.

\begin{figure}
  \centering
  \includegraphics[width=\linewidth]{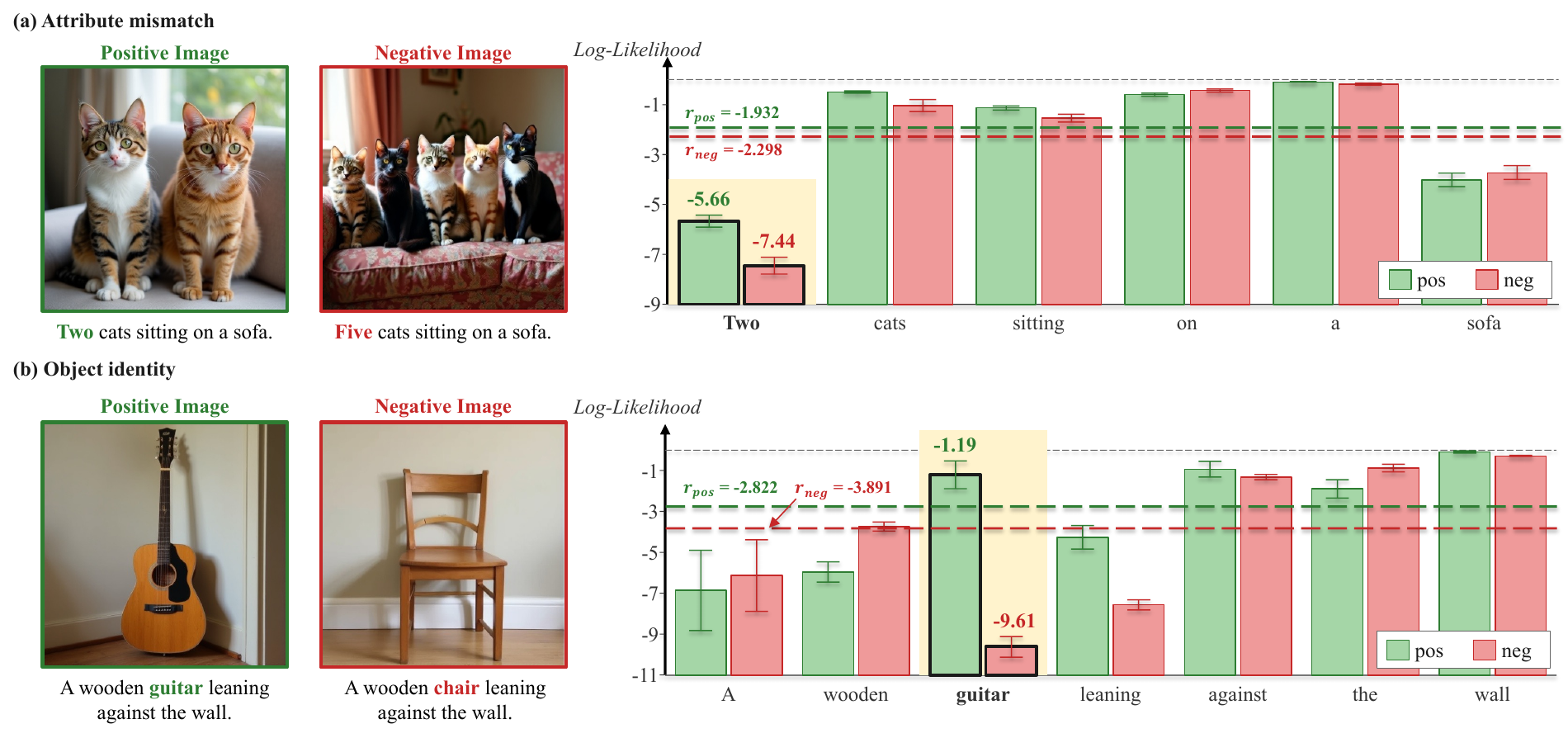}
  \caption{
    \textbf{Token-level semantic sensitivity of \method.}
    Positive and negative images are evaluated using the same positive prompt. 
    For attribute mismatch, instantiated as a counting error, 
    the negative image mainly lowers the likelihood of ``Two''; 
    for object identity mismatch, replacing the guitar with a chair sharply lowers the likelihood of ``guitar''. 
    Bars show image-conditioned prompt-token log-likelihoods with error bars calculated over four pairs, 
    and dashed lines show the resulting sequence-level reward value, i.e., \method.
}
  \label{fig:per_token_reward}
    \vspace{-12pt}
\end{figure}

\subsection{Analysis of \method}
\label{sec:analysis}

In this section, we analyze three questions:
(1) whether the language prior affects the accuracy of the reward signal;
(2) whether token-level likelihoods are sensitive to visual misalignment;
(3) whether the resulting scalar reward can reliably distinguish samples within the same prompt group.

\noindent \textbf{Language-prior cancellation.}
A natural concern is that~\autoref{eq:method_reward} contains the MLLM's language prior over the prompt, since some words are easier to
predict regardless of the image. 
One possible correction is to subtract a text-only likelihood term, 
yielding a PMI-style reward $\log p(x \mid y) - \log p(x)$. 
However, since we focus on using group-relative reinforcement learning methods,
this correction is unnecessary. 
For a fixed prompt, the text-only likelihood is shared by all generated images in the same group,
and therefore cancels out when computing the group-relative advantage. 
Thus, PMI normalization may affect absolute reward calibration across prompts, 
but it does not provide additional optimization signal within each prompt group.

\noindent \textbf{Token-level semantic sensitivity.} We further examine whether the semantic spectrum responds to local visual errors.
We construct positive and negative image pairs for two
common prompt-following failures, attribute mismatch and object identity
mismatch. The attribute mismatch is instantiated as a counting error,
where the prompt asks for two cats but the negative image contains five.
For each pair, we evaluate both images using the original positive prompt.~\autoref{fig:per_token_reward} shows that the likelihood drop is concentrated on the mismatched semantic word. 
The counting mismatch mainly lowers the likelihood of ``Two'', 
while replacing the guitar with a chair sharply lowers the likelihood of ``guitar''. 
The sequence-level reward, shown by the dashed lines, also remains higher for the positive images. 
This indicates that token-level prompt likelihoods are sensitive to targeted semantic errors, 
and their average, i.e., the calculation used by \method, still provides a reliable scalar reward.

\begin{figure}
  \centering
  \includegraphics[width=\linewidth]{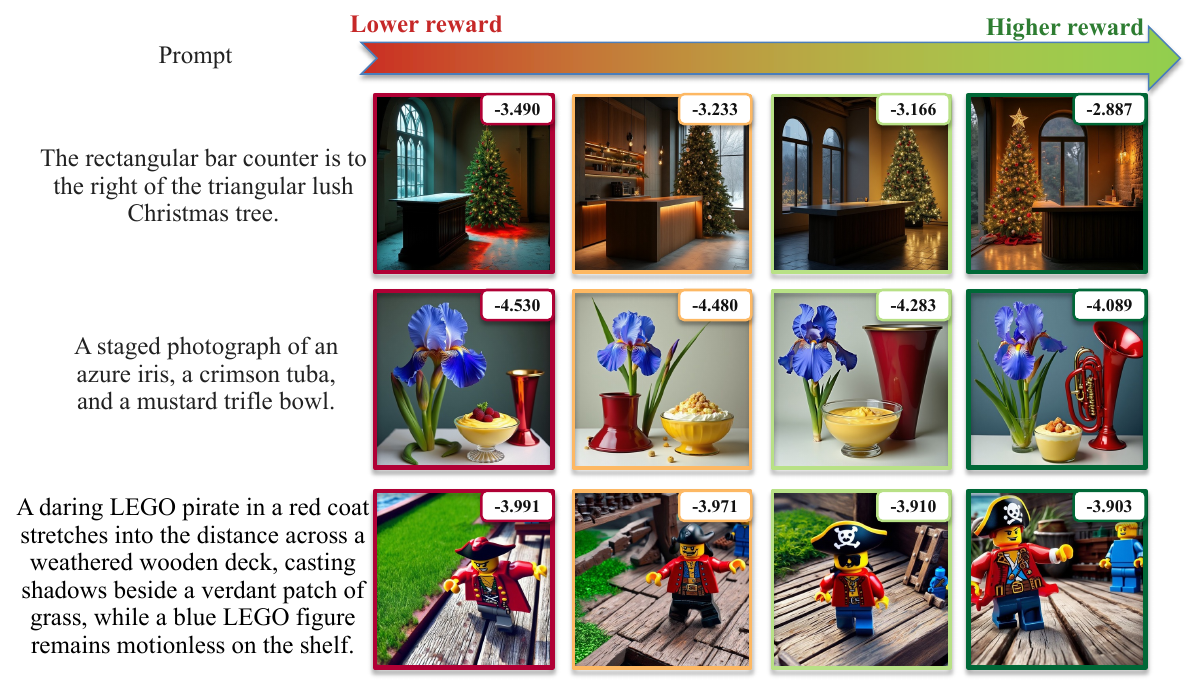}
  \vspace{-2mm}
  \caption{\textbf{The visual interpretation of \method.} 
  The reward ranking is consistent with the visual quality ranking.}
  \label{fig:reward_ranking}
  \vspace{-10pt}
\end{figure}

\noindent \textbf{Reward ranking reliability.}  
What matters for group-relative RL is whether the reward correctly orders
different rollouts from the same prompt. As shown in~\autoref{fig:reward_ranking}, \method assigns higher rewards to images
that better satisfy the prompt and lower rewards to images with missing
objects, wrong attributes, or incorrect spatial relations. This indicates
that the image-conditioned prompt likelihood provides a reliable
group-wise ranking signal for text-to-image RL.

\section{Experiments}
\label{sec:experiments}

\subsection{Experimental Setup}
\label{sec:implementation}
We use AWM as the default reinforcement learning algorithm, following the
optimizer ablation in~\autoref{sec:ablation_algo}, where AWM achieves the
best overall performance among the tested RL algorithms.
Our policy model is BAGEL~\cite{bagel}, a native AR-Diffusion unified multimodal model pretrained for both image understanding and image generation.
Unless otherwise specified, \method uses Qwen3-VL-30B-A3B~\cite{qwen3vl} as the
reward MLLM backbone, while Self-\method uses BAGEL's own understanding
branch for reward computation.

\noindent \textbf{Training settings.} 
We use $T=16$ sampling steps during training and stochastically sample 6 steps from the first 10 denoising steps for training. 
Each training step uses $32$ prompts, and the group size is $G=16$. The
training image resolution is $512 \times 512$. We use a classifier-free
guidance scale of $4$ and a timestep shift of $3$. The learning rate is 1e-4.
We trained on 32 A100 GPUs with batch size of 2 and gradient accumulation of 8.
The total training steps are 380.  Other hyperparameters
follow the official AWM implementation. We train on AlphaGRPO20k~\citep{alphagrpo}.

\begin{table*}[t]
\centering
\tablestyle{3pt}{1.2}
\caption{\textbf{Results on Text-to-Image benchmarks}. S and L denote Short and Long prompts, respectively.
For WISE, the second score denotes evaluation with CoT. 
AlphaGRPO is trained on the reasoning text-to-image generation task. 
\textbf{Bold} indicates the best performance.
The results of BAGEL are reproduced.}
\resizebox{\textwidth}{!}{%
\begin{tabular}{lcccccccccccc}
\toprule[1.5pt]
\multirow{3}{*}{\textbf{Model}} & \multicolumn{8}{c}{\textbf{TIIF Bench} ↑} & \multirow{2}{*}{\textbf{WISE} ↑} & \multirow{2}{*}{\textbf{DPGBench} ↑} & \multirow{2}{*}{\textbf{GenEval} ↑} & \multirow{2}{*}{\textbf{GenEval2} ↑} \\
& \multicolumn{2}{c}{Basic} & \multicolumn{2}{c}{Advanced} & \multicolumn{2}{c}{Designer} & \multicolumn{2}{c}{{Overall}} & & & & \\
& S & L & S & L & S & L & S & L & Overall & Score & Score & {Soft TIFA}$_{\mathbf{GM}}$ \\
\midrule
\rowcolor{gray!10}
\multicolumn{13}{l}{\textit{Generation Only Models}} \\
SD3 Medium~\cite{sd3} & 78.3 & 77.8 & 61.5 & 59.6 & 63.2 & 67.3 & 64.8 & 64.8 & 0.4 & 84.1 & 74.0 & 21.3 \\
FLUX.1 dev~\cite{flux} & 83.1 & 78.7 & 65.8 & 68.5 & 70.7 & 71.5 & 71.1 & 71.8 & 0.5 & 83.8 & 82.0 & 21.1 \\
\midrule
\rowcolor{gray!10}
\multicolumn{13}{l}{\textit{Unified Multimodal Models}} \\
Show-o~\cite{showo}       & 73.1 & 75.8 & 55.0 & 50.9 & 53.7 & 50.4 & 59.7 & 58.9 & 0.35 & - & 69.0 & - \\
JanusPro~\cite{januspro}  & 79.3 & 78.3 & 59.7 & 58.8 & 65.8 & 60.3 & 66.5 & 65.0 & 0.35 & 84.2 & 80.0 & - \\
\midrule
\rowcolor{gray!10}
\multicolumn{13}{l}{\textit{Inference on 512 Resolution}} \\
BAGEL & 81.7 & 86.1 & 73.7 & 77.6 & 84.7  & 82.1 & 75.2 & 78.6 & - & 85.0 & 84.0 & 23.6 \\
AlphaGRPO~\cite{alphagrpo} & 85.5 & 84.2 & 77.4 & 78.9 & 84.3 & 86.6 & 78.9 & 79.5 & - & 86.0 & 85.0 & - \\
\rowcolor{myblue}
\textbf{\method} & 87.7 & 87.4 & \textbf{84.9} & 85.4 & 88.1 & \textbf{90.3} & \textbf{85.2} & 84.8 & - & \textbf{88.1} & 88.3 & 31.8\\
\rowcolor{myblue}
\textbf{Self-\method} & \textbf{89.7} & \textbf{87.6} & 83.7 & \textbf{85.7} & \textbf{90.3} & 86.2 & 85.1 & \textbf{84.3} & - & 87.7 & \textbf{89.5} & \textbf{33.9}
\\
\midrule
\rowcolor{gray!10}
\multicolumn{13}{l}{\textit{Inference on 1024 Resolution}} \\
BAGEL & 83.4 & 83.7 & 75.2 & 76.7 & 79.8 & 73.5 & 76.4 & 76.2 & 0.53/0.70 & 85.1 & 86.6 & 23.5 \\
AlphaGRPO~\cite{alphagrpo} & 84.8 & 85.9 & 79.9 & 78.4 & 80.2 & 80.2 & 78.7 & 79.5 & 0.50/0.69 & 85.9 & 87.4 & - \\
\rowcolor{myblue}
\textbf{\method} & \textbf{88.4} & 87.3 & \textbf{83.2} & 81.8 & 82.5 & 81.0 & \textbf{83.3} & 81.7 & \textbf{0.53}/0.74 & 86.5 & 88.3 & 32.6 \\
\rowcolor{myblue}
\textbf{Self-\method} & 87.4 & \textbf{87.6} & 81.5 & \textbf{83.2} & \textbf{87.7} & \textbf{84.7} & 82.7 & \textbf{82.5} & 0.52/\textbf{0.76} & \textbf{87.0} & \textbf{89.8} & \textbf{34.3} \\
\bottomrule[1.5pt]
\end{tabular}
}
\label{tab:main_result}
\vspace{-5pt}
\end{table*}

\subsection{Main Results}
\label{sec:main_results}
We use our \method as the proxy reward during training and evaluate the real improvement of generation performance on five out-of-distribution benchmarks.
All downstream benchmarks use different evaluation protocols and out-of-distribution test sets: 
GenEval~\cite{geneval}, TIIF-Bench~\cite{tiif}, DPG-Bench~\cite{dpg}, Geneval2~\cite{kamath2025geneval}, and WISE~\cite{wise}.
We compare against the state-of-the-art generation-only models, 
including SD3 Medium~\cite{sd3} and FLUX.1 dev~\cite{flux}, and {unified multimodal models} such as 
Show-o~\cite{showo}, JanusPro~\cite{januspro}, and our baseline BAGEL~\cite{bagel}. 
We compare with AlphaGRPO~\cite{alphagrpo}, which also applies the RL training on Bagel with the proposed MLLM-derived DVReward.

\autoref{tab:main_result} compares \method and
Self-\method with representative text-to-image generators and prior
MLLM-derived reward training methods. Overall, both \method and
Self-\method achieve comparable
performance and consistently improve over the BAGEL baseline and AlphaGRPO
across all evaluated benchmarks. 
Specifically, at 512 resolution, \method improves BAGEL on TIIF-Bench overall
short/long prompts by +10.0/+6.2, while
Self-\method achieves a comparable 85.1/84.3 and further brings a 5.5 gain on GenEval. 
Compared with AlphaGRPO, \method improves
TIIF-Bench by +6.3/+5.3 on short/long prompts and +3.3 on GenEval, while
Self-\method improves them by +6.2/+4.8 and +4.5, respectively.

Consistent with AlphaGRPO~\cite{alphagrpo}, although training is performed at 512 resolution, 
the improvements also transfer to 1024-resolution inference. 
Specifically, Self-\method reaches 89.8 on GenEval and 34.3 on GenEval2, outperforming both BAGEL and
AlphaGRPO. More importantly, Self-\method achieves 0.76 on the knowledge-grounded benchmark, WISE.
These results suggest that \method provides a more effective supervision signal than
decompositional verification for text-to-image RL.

\begin{table}[t]
    \scriptsize
    \centering
    \tablestyle{4pt}{1.2}
    \caption{ \textbf{Effect of different reward MLLM backbones.}
    Unless otherwise specified, reward MLLMs all use Instruct models.
    \textbf{Bold} indicates the best performance. The second best is \underline{underlined}.}
    \label{tab:reward_mllm}
    \begin{tabular}{llccc}
    \toprule[1.5pt]
    \textbf{Base Model} & \textbf{Reward MLLM} & \textbf{GenEval$\uparrow$} & \textbf{TIIF-Short$\uparrow$} & \textbf{TIIF-Long$\uparrow$} \\
    \midrule
    SD3.5-M & -                       & 79.8 & 74.0 & 73.2\\
    SD3.5-M & Qwen3-VL-8B             & 85.7 & 84.4 & 84.6 \\
    \midrule
    BAGEL & -                       & 84.0 & 75.2 & 78.6 \\
    BAGEL & Gemma3-4B      & 83.6 & 83.1 & 83.0 \\
    BAGEL & Gemma3-12B-Pretrain     & \underline{88.7} & 84.3 & 82.6 \\
    BAGEL & Gemma3-12B     & 87.5 & 82.0 & 82.6 \\
    BAGEL & InternVL3.5-8B          & 84.5 & 80.4 & 80.0 \\
    BAGEL & InternVL3.5-14B         & 84.9 & 78.5 & 79.9 \\
    BAGEL & Qwen3-VL-8B             & 87.8 & 85.0 & 83.8 \\
    \rowcolor{gray!10}
    BAGEL & Qwen3-VL-30B-A3B        & 88.3 & \textbf{85.2} & \textbf{84.8} \\
    BAGEL & Qwen3-VL-235B-A22B      & 86.8 & 84.3 & 83.4 \\
    \rowcolor{gray!10}
    BAGEL & BAGEL                   & \textbf{89.5} & \underline{85.1} & \underline{84.3} \\
    \bottomrule[1.5pt]
    \end{tabular}
\end{table}

\begin{table}[t]
    \scriptsize
    \centering
    \tablestyle{4pt}{1.2}
    \caption{\textbf{Comparison of different RL algorithms and reward models.}
    When trained with Self-\method, AWM achieves the best downstream performance.
    Self-\method further outperforms prior reward models under comparable RL training.
    \textbf{Bold} indicates the best performance. }
    \label{tab:rl_algo}
    \begin{tabular}{lcccc}
    \toprule[1.5pt]
    \textbf{RL Algorithm} & \textbf{Reward Model} & \textbf{GenEval$\uparrow$} & \textbf{TIIF-Short$\uparrow$} & \textbf{TIIF-Long$\uparrow$} \\
    \midrule
    -         & - & 84.0 & 75.2 & 78.6 \\
    AlphaGRPO & HPSv3          & 83.4  & 78.5 & 77.1 \\
    AlphaGRPO & UnifiedReward  & 83.7 & 79.2 & 77.3  \\
    AlphaGRPO & VIEScore       & 81.7 & 79.1 & 77.9  \\
    AlphaGRPO & DVReward       & 85.0 & 78.9 & 79.5 \\
    AWM       & DVReward       & 88.0 & 83.3 & 83.7 \\
    \midrule
    FlowGRPO     & Self-\method & 85.3 & 82.1 & 79.8 \\
    DiffusionNFT & Self-\method & 89.1 & 84.8 & 83.5 \\
    \rowcolor{gray!10}
    AWM          & Self-\method & \textbf{89.5} & \textbf{85.1} & \textbf{84.3} \\
    \bottomrule[1.5pt]
    \end{tabular}
\end{table}

\subsection{Effect of reward MLLM backbone}
\label{sec:reward_backbone}
We next study the effect of different reward MLLM backbones in \method, covering different families and scales.~\autoref{tab:reward_mllm} reports the experimental results.

\noindent \textbf{Consistent improvement of \method across diffusion models and reward MLLM families.}
On both SD3.5-M~\cite{sd3} and BAGEL, applying \method improves the baseline, 
highlighting the architecture-agnostic nature of the caption-likelihood reward. 
On BAGEL, different reward MLLMs from different families, including Gemma3~\cite{gemma3}, InternVL3.5~\cite{wang2025internvl3}, and Qwen3-VL~\cite{qwen3vl}, 
all bring improvements over the baseline on TIIF-Bench. 
This indicates that \method can extract useful reward signals from diverse MLLM backbones.

\noindent \textbf{Self-\method shows that reward-policy alignment can rival external scale.}
Using BAGEL's own understanding branch as the reward MLLM, Self-\method
achieves the best result among similar-scale reward models, matches the
strongest 30B-class external reward, and outperforms 30x larger MLLMs, i.e., Qwen3-VL-235B-A22B~\cite{qwen3vl}.
This suggests that reward quality is not determined by scale alone.
We attribute the success of Self-\method to its
distributional alignment with the policy: Self-\method shares the
tokenizer, vision encoder, and pretraining distribution with the
generator, so the way it reads generated images is naturally matched to
the distribution from which the policy samples them.

\noindent \textbf{Reward MLLM scale helps, but is not monotonic.}
Within Qwen3-VL, scaling from 8B to 30B improves all reported metrics, 
but further scaling to 235B leads to a clear drop. 
Within the Gemma3 family, scaling from 4B to 12B improves GenEval, but does not consistently improve TIIF-Bench. 
We hypothesize that larger MLLMs may devote more capacity to 
advanced vision-language reasoning and instruction-following tasks, 
while the more fundamental image-conditioned captioning ability needed by \method may already saturate at a moderate scale. 
These results suggest that a 30B MLLM is sufficient to unlock most of the benefit of \method.

\noindent \textbf{Pretraining-stage MLLMs can be better reward backbones for \method.}
Gemma3-12B-Pretrain consistently outperforms Gemma3-12B-Instruct across three evaluated benchmarks.
We hypothesize that pretraining learns large-scale image-captioning and image-text alignment objectives that closely match
our reward function, while instruction tuning emphasizes advanced visual tasks and multi-task instruction following
that are less directly needed by \method. 

\begin{table*}[t]
    \scriptsize
    \centering
    \begin{minipage}[t]{0.48\textwidth}
        \centering
        \captionof{table}{\textbf{Ablation of Reward function.}
        Scalar Scoring asks the MLLM to rate the image from 1 to 5.
        VQA-Score asks ``Does this figure show \texttt{\{caption\}}?'' and
        uses $P(\text{yes})$ as the reward.}
        \label{tab:reward_form}
        \resizebox{\linewidth}{!}{%
        \begin{tabular}{lccc}
        \toprule[1.5pt]
        \textbf{Reward function} & \textbf{ GenEval$\uparrow$} & \textbf{TIIF-S$\uparrow$} & \textbf{TIIF-L$\uparrow$} \\
        \midrule
        - & 84.0 & 75.2 & 78.6 \\
        Scalar Scoring (1-5) & 77.7 & 67.5 & 76.0 \\
        VQA-Score ($P(\text{yes})$) & 83.1 & 77.4 & 78.9 \\
        \rowcolor{gray!10}
        Prompt Likelihood & \textbf{89.5} & \textbf{85.1} & \textbf{84.3} \\
        \bottomrule[1.5pt]
        \end{tabular}%
        }
    \end{minipage}
    \hfill
    \begin{minipage}[t]{0.47\textwidth}
        \centering
        \captionof{table}{\textbf{Ablation of advantage aggregation.}
        All variants produce one image-level advantage; token-level variants
        differ in normalization before aggregation.}
        \label{tab:token_adv}
        \resizebox{\linewidth}{!}{%
        \begin{tabular}{lccc}
        \toprule[1.5pt]
        \textbf{Method} & \textbf{GenEval$\uparrow$} & \textbf{TIIF-S$\uparrow$} & \textbf{TIIF-L$\uparrow$} \\
        \midrule
        \rowcolor{gray!10}
        Sequence-level Adv. & \textbf{89.5} & \textbf{85.1} & 84.3 \\
        Token-level Adv. & 88.4 & 84.9 & \textbf{85.1} \\
        \quad + Group std & 88.5 & 84.0 & 84.8 \\
        \quad + Per-token std & 88.8 & 84.1 & 82.1 \\
        \bottomrule[1.5pt]
        \end{tabular}%
        }
    \end{minipage}
\end{table*}

\begin{table*}[t]
    \centering
    \tablestyle{4pt}{1.2}
    \caption{\textbf{Generalization across generator backbones.}
    Z-Image and Qwen-Image use DiffusionNFT for RL post-training.
    Avg reports the unweighted mean performance across benchmarks,excluding GenEval2.
    Bold indicates the best score within each generator; dashes denote unreported results.}
    \label{tab:additional_generators}
    \begin{tabular}{@{}lcccccc@{}}
    \toprule
    \textbf{Model} & \textbf{GenEval} & \textbf{TIIF-S} & \textbf{TIIF-L} & \textbf{DPG} & \textbf{GenEval2} & \textbf{Avg} \\
    \midrule
    Z-Image & 81.94 & 86.61 & 86.03 & 86.54 & 30.59 & 85.28 \\
    \quad + \method & \textbf{90.99} & \textbf{92.30} & \textbf{93.10} & \textbf{88.85} & \textbf{49.69} & \textbf{91.31} \\
    \midrule
    Qwen-Image & 87.94 & 82.38 & 78.81 & 86.10 & 32.35 & 83.81 \\
    \quad + \method & \textbf{90.45} & \textbf{92.54} & \textbf{90.02} & \textbf{89.27} & \textbf{60.09} & \textbf{90.57} \\
    \midrule
    OmniGen2 & 81.22 & 77.35 & 77.42 & 82.82 & -- & 79.70 \\
    \quad + \method & 87.65 & \textbf{82.16} & 77.87 & \textbf{87.21} & -- & 83.72 \\
    \quad + Self-\method & \textbf{89.05} & 82.11 & \textbf{81.31} & 86.92 & -- & \textbf{84.85} \\
    \bottomrule
    \end{tabular}
\end{table*}

\subsection{Ablation Study}
\label{sec:ablation}

\noindent \textbf{The effect of RL algorithm.}
\label{sec:ablation_algo}
We compare three image generation reinforcement learning methods in the Self-\method setting:
FlowGRPO~\citep{flowgrpo}, AWM~\citep{awm}, and DiffusionNFT~\citep{diffusionnft}.~\autoref{tab:rl_algo} shows that AWM achieves the best overall downstream performance, outperforming the SDE-based FlowGRPO and DiffusionNFT.
We therefore apply AWM in the main experiments.

\begin{figure*}[t]
  \centering
  \includegraphics[width=\linewidth]{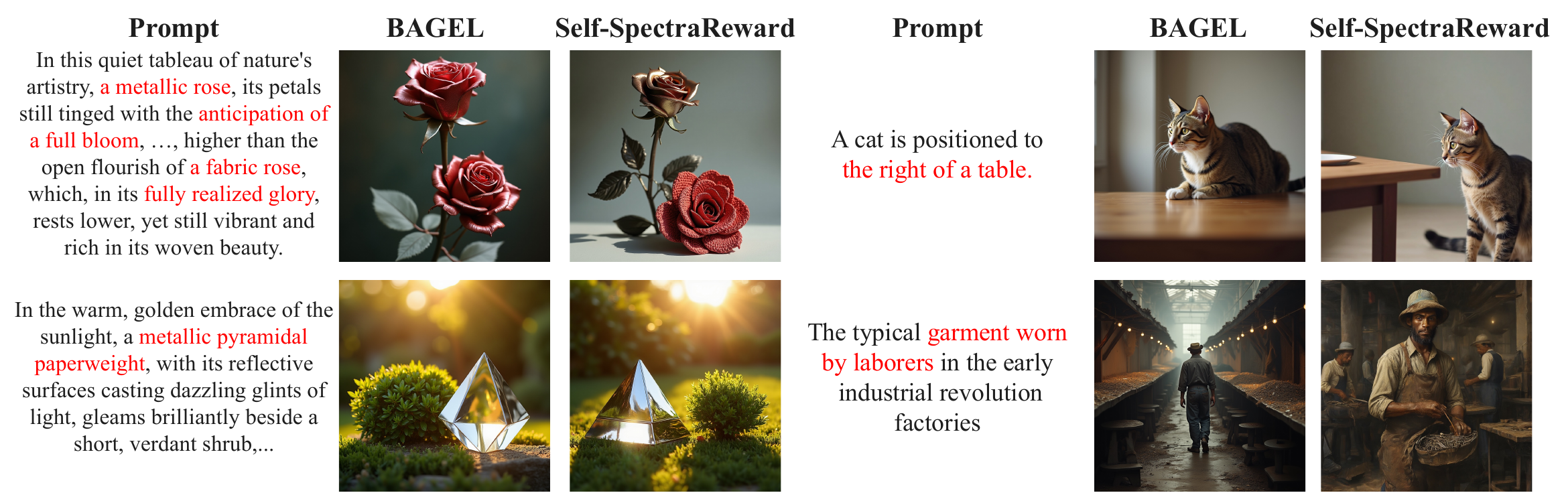}
  \caption{\textbf{Qualitative comparison.}}
  \label{fig:qualitative}
  \vspace{-10pt}
\end{figure*}

\noindent \textbf{Discussion of reward function.}
\label{sec:ablation_reward_form}
We compare \method with two alternative MLLM-based reward functions for measuring image-text alignment.
As illustrated in~\autoref{fig:qualitative}, the first is scalar scoring,
which directly asks the MLLM to rate image-text alignment with a $1$--$5$ score. 
The second is VQA-Score, which asks a yes/no question such as ``Does this figure show \texttt{\{text\}}?'' and uses the
probability of the ``yes'' token as the reward.~\autoref{tab:reward_form} shows that the reward function has a large impact on RL performance.
Scalar scoring substantially degrades all benchmarks, including a $6.3$ drop on GenEval compared with the baseline, 
while VQA-Score brings minor gains on TIIF-Bench but fails to improve GenEval.
In contrast, the image-conditioned prompt likelihood used by \method
achieves the best performance across all metrics, showing that
likelihood-based reward is a more effective way to leverage MLLMs for
image-generation RL.

\noindent \textbf{Effect of advantage aggregation.}
\label{sec:ablation_token}
\autoref{tab:token_adv} compares the default sequence-level advantage with the token-level variant defined in~\autoref{sec:reward_def}, using global, group, and per-token standard deviations.
The global-std token-level variant performs the best compared to other token-level variants and it improves TIIF-Long from 84.3 to 85.1, 
but reduces GenEval from 89.5 to 88.4 and TIIF-Short from 85.1 to 84.9.
The long-prompt gain is also sensitive to normalization: per-token std yields 82.1 on TIIF-Long.
These results show no consistent benefit from token-level aggregation across the evaluated benchmarks and normalization schemes.
We therefore retain sequence-level advantage as the default; evaluation on substantially longer prompts remains future work.

\subsection{Generalization across Generator Backbones}
\label{sec:additional_generators}
Using DiffusionNFT, we further optimize Z-Image and Qwen-Image with \method.
Both models improve on all five evaluated metrics, as shown in~\autoref{tab:additional_generators}.
Z-Image improves by 9.05 points on GenEval and 19.10 points on GenEval2.
Qwen-Image improves by 10.16 and 11.21 points on TIIF-Short and TIIF-Long, respectively, and by 27.74 points on GenEval2.

\noindent \textbf{Self-rewarding beyond BAGEL.}
On OmniGen2, external \method uses Qwen3-VL-30B-A3B, while Self-\method uses the model's own Qwen2.5-VL-3B understanding branch.
The four-metric mean Avg rises from 79.70 to 83.72 with external \method and to 84.85 with Self-\method.
Self-\method scores higher on GenEval and TIIF-Long, while external \method scores higher on TIIF-Short and DPG.
These results support self-rewarding beyond BAGEL and are consistent with our reward-policy alignment hypothesis.

\section{Conclusion}
\label{sec:conclusion}
We introduced \method, a training-free reward function that turns
pretrained MLLMs into off-the-shelf reward models for image-generation
reinforcement learning by measuring image-conditioned prompt likelihood. 
Instead of relying on massive preference labels, reward-model fine-tuning,
\method aggregates the token-level semantic spectrum induced by an MLLM into a simple and effective reward. 
We further introduced Self-\method for unified multimodal models, 
where the policy's own understanding branch serves as the reward model for its generation branch, 
forming a closed-loop self-improving framework without external reward models. 
Across generator backbones, RL algorithms, reward MLLM backbones, and out-of-distribution benchmarks, 
\method significantly and consistently improves text-to-image generation and 
outperforms prior MLLM-derived reward training methods. 
The results also show that larger reward MLLMs are not always better, 
while a well-aligned self-reward can match or surpass much larger external reward models. 
These findings suggest that reward function and reward-policy alignment are key factors for
reinforcement learning in unified multimodal generation.

\bibliographystyle{plainnat}
\bibliography{ref}

\newpage
\appendix
\newpage
\appendix
\setcounter{page}{1}

\section*{\textbf{Appendix}}

\renewcommand{\subsectionautorefname}{Appendix}
\renewcommand{\sectionautorefname}{Appendix}

The outline of the Appendix is as follows.
\begin{itemize}[leftmargin=*, itemsep=2pt, topsep=2pt]
\item \ref{sec:app_limitations} Limitations and Future Work. We discuss the limitations of prompt-likelihood rewards, including dependence on MLLM visual reasoning, implicit physical and commonsense implications, and complementary reward signals.
\item \ref{sec:additional_implementation_details} Additional Implementation Details, including the reward prompt format, EOS masking, and the use of VAE features in Self-\method on BAGEL.
\item \ref{sec:additional_experimental_results} Additional Experimental Results, including EOS and VAE feature ablations, the full AlphaGRPO SRR comparison, and a comparison with a task-specific GenEval reward.

\item \ref{app:detailed_benchmark} Detailed Benchmark Results, including category-level results on GenEval, TIIF-Bench, DPG-Bench, and WISE.
\item \ref{sec:more_visualizations} More Visualizations, including qualitative comparisons on spatial relations, counting, attribute binding, relative size, and long-prompt composition.
\item \ref{sec:broader_impacts} Broader Impacts.
\end{itemize}

\section{Limitations and future work}
\label{sec:app_limitations}
\method relies on the image-conditioned prompt likelihood of pretrained MLLMs, 
so its reward quality is bounded by the visual understanding and reasoning ability of the chosen reward backbone. 
Since the likelihood is computed only over the input prompt, the reward mainly captures explicit semantic alignment between generated images and prompts, 
and may under-emphasize implicit visual implications that are not directly expressed in the text. 
For example, a prompt such as ``hot coffee'' may imply steam through physical and commonsense reasoning, 
but such implications may be weakly reflected in the likelihood of the original prompt tokens. 
This limitation may be alleviated by using stronger MLLM backbones with rich world knowledge, 
or by extending \method to reasoning text-to-image generation, where intermediate reasoning can make such implicit requirements explicit. 
In addition, \method does not directly optimize criteria such as aesthetics and safety, 
which could be addressed by combining it with complementary specialized reward models. 

\section{Additional Implementation Details}
\label{sec:additional_implementation_details}
When using \method, we feed the generated image as the visual condition
and evaluate the original prompt with teacher forcing. We do not add any
general prefix to ask the reward MLLM to describe the image. Although
such a prefix can change absolute likelihood values, it does not
substantially affect the relative ranking within each prompt group, which
is the quantity used by group-relative RL. Therefore, we refrain from
using a prefix for all reward MLLMs by default. The only exception is the
InternVL3.5 family, where we empirically find that adding the prefix
``Describe the image'' gives more stable reward values.

We exclude the [EOS] token from the \method computation. The [EOS]
likelihood is often dominated by sequence termination rather than
image-text alignment, and it can disproportionately affect short prompts
because it contributes a large fraction of the averaged token likelihood.
The ablation in~\autoref{sec:ablation_eos} shows that masking [EOS]
improves GenEval while keeping TIIF-Bench performance comparable.

For Self-\method on BAGEL, we use the model's own understanding branch to
score generated images. BAGEL contains both semantic visual features and
VAE features. By default, we include the VAE feature in Self-\method
because it gives better average performance in our ablation. We discuss
the effect of VAE inputs in~\autoref{sec:ablation_vae}.

\section{Additional Experimental Results}
\label{sec:additional_experimental_results}

\paragraph{Effect of EOS token removal.}
\label{sec:ablation_eos}
When inspecting the per-token likelihood distribution, 
we find that the EOS token often receives much lower likelihood than regular prompt tokens. 
Because \method averages token log-likelihoods, this unusually low-likelihood token can significantly shift the final reward, especially for short prompts. 
However, EOS only indicates sequence termination and does not carry semantic information about image-text alignment. 
We therefore mask the EOS token when computing \method.

\paragraph{Effect of VAE feature in the reward calculation on BAGEL.}
\label{sec:ablation_vae}
BAGEL uses separate encoders for generation and understanding, 
i.e., ViT for understanding and VAE for generation, and can optionally include the VAE feature when conducting understanding tasks.
In Self-\method, we use BAGEL's understanding backbone to provide reward signals for generation.
Here, we ablate the effect of using the VAE feature in Self-\method.~\autoref{tab:vae} shows that including the VAE feature obtains better performance on average, 
and the improvement is more significant on the TIIF-Long benchmark.

\paragraph{Comparison with AlphaGRPO's self-reflective refinement.}
\label{app:alphagrpo_srr}
The main comparison uses AlphaGRPO's reasoning text-to-image (RT2I) variant, which more closely matches our generation setting.
\autoref{tab:alphagrpo_srr} additionally reports its self-reflective refinement (SRR) configurations at 512 resolution, with and without inference-time SRR.
\method and Self-\method use a single generation pass, while inference-time SRR adds reflection and image refinement after the initial generation, incurring additional inference cost.
The comparison should therefore be interpreted with this difference in inference budget in mind.
\method scores higher than AlphaGRPO with inference SRR on all four reported metrics.
Self-\method scores higher on three metrics, including GenEval (89.5 vs.\ 88.2), while its DPG score is slightly lower (87.7 vs.\ 87.9).

\begin{table}[t]
    \centering
    \tablestyle{4pt}{1.2}
    \caption{\textbf{Full comparison with AlphaGRPO's SRR configurations at 512 resolution.}
    Inference-time SRR uses additional reflection and image refinement beyond the initial generation.
    Bold indicates the best score in each column.}
    \label{tab:alphagrpo_srr}
    \begin{tabular}{lccccc}
    \toprule
    \textbf{Method} & \textbf{Inference SRR} & \textbf{TIIF-S} & \textbf{TIIF-L} & \textbf{DPG} & \textbf{GenEval} \\
    \midrule
    \method & No & \textbf{85.2} & \textbf{84.8} & \textbf{88.1} & 88.3 \\
    Self-\method & No & 85.1 & 84.3 & 87.7 & \textbf{89.5} \\
    AlphaGRPO (SRR)~\cite{alphagrpo} & No & 79.1 & 79.5 & 86.3 & 84.2 \\
    AlphaGRPO (SRR)~\cite{alphagrpo} & Yes & 83.9 & 83.2 & 87.9 & 88.2 \\
    \bottomrule
    \end{tabular}
\end{table}

\paragraph{Task-specific rewards and cross-benchmark generalization.}
\label{app:task_specific_reward}
We compare SD3.5 optimized with the task-specific GenEval score against SD3.5 optimized with \method in~\autoref{tab:geneval_reward}.
The task-specific reward achieves a higher score on its target benchmark, GenEval (92.8 vs.\ 85.7).
\method obtains higher scores on all four remaining metrics, including TIIF-Long (84.6 vs.\ 75.3) and GenEval2 (30.6 vs.\ 23.7).
These results illustrate the distinction between optimizing a target evaluation metric and obtaining gains across different evaluation protocols, supporting our focus on cross-benchmark generalization.

\begin{table}[t]
    \centering
    \tablestyle{4pt}{1.2}
    \caption{\textbf{Task-specific GenEval reward versus \method on SD3.5.}
    The GenEval-score reward directly targets GenEval; the remaining columns evaluate transfer to other metrics.
    Bold indicates the best score in each column.}
    \label{tab:geneval_reward}
    \begin{tabular}{lccccc}
    \toprule
    \textbf{Training reward} & \textbf{GenEval} & \textbf{TIIF-S} & \textbf{TIIF-L} & \textbf{DPG} & \textbf{GenEval2} \\
    \midrule
    None (SD3.5 baseline) & 79.8 & 74.0 & 73.2 & 84.6 & 18.0 \\
    GenEval-score & \textbf{92.8} & 78.8 & 75.3 & 85.2 & 23.7 \\
    \method & 85.7 & \textbf{84.4} & \textbf{84.6} & \textbf{89.2} & \textbf{30.6} \\
    \bottomrule
    \end{tabular}
\end{table}

\section{Detailed Benchmark Results}
\label{app:detailed_benchmark}

This section provides the full category-level benchmark results that complement the aggregate numbers in~\autoref{tab:main_result}. 
We include detailed breakdowns on GenEval, TIIF-Bench, DPG-Bench, and WISE to 
examine whether the gains from \method are concentrated in a single score or hold across different types of text-to-image requirements. 
Overall, the detailed results show that both \method and Self-\method improve the BAGEL baseline across compositional generation, fine-grained instruction following and knowledge-grounded generation. 
The improvements also remain visible when the same 512-resolution trained model is evaluated at 1024 resolution, 
suggesting that the learned reward signal transfers beyond the training sampling resolution.

\begin{table}[t]
    \scriptsize
    \centering
    \tablestyle{4pt}{1.2}
    \caption{\textbf{The effect of including the EOS token in the reward calculation.}
    }
    \label{tab:eos}
    \begin{tabular}{cccc}
    \toprule
    \textbf{Include EOS token} & \textbf{GenEval$\uparrow$} & \textbf{TIIF-Short$\uparrow$} & \textbf{TIIF-Long$\uparrow$} \\
    \midrule
    Yes            & 88.5 & \textbf{85.5} & \textbf{84.3} \\
    \rowcolor{gray!10}
    No           & \textbf{89.5} & 85.1 & \textbf{84.3} \\
    \bottomrule
    \end{tabular}
\end{table}

\begin{table}[t]
    \scriptsize
    \centering
    \tablestyle{4pt}{1.2}
    \caption{\textbf{Effect of VAE features in Self-\method reward computation.}
        BAGEL provides two visual inputs to its understanding branch, 
        ViT features from the semantic encoder and VAE features from the generation encoder. 
        Using both inputs improves GenEval and TIIF-Short.
    }
    \label{tab:vae}
    \begin{tabular}{cccc}
    \toprule
    \textbf{Use VAE input} & \textbf{GenEval$\uparrow$} & \textbf{TIIF-Short$\uparrow$} & \textbf{TIIF-Long$\uparrow$} \\
    \midrule
    No             & 87.8 & 83.6 & \textbf{84.7}  \\
    \rowcolor{gray!10}
    Yes           & \textbf{89.5} & \textbf{85.1} & {84.3} \\
    \bottomrule
    \end{tabular}
\end{table}

\paragraph{GenEval.}~\autoref{tab:geneval} reports the complete GenEval breakdown. 
Compared with BAGEL and AlphaGRPO, \method and Self-\method improve the overall score at both 512 and 1024 inference resolutions. 
The largest improvements appear in the more compositional categories, including counting, spatial position and color-attribute binding, 
where successful generation requires not only object presence but also correct attribute binding to objects.
Self-\method achieves the best overall score at both resolutions, indicating that using the policy model's own understanding branch 
as the reward model can provide a particularly well-aligned signal for structured prompt following.
\definecolor{gray10}{gray}{0.9}

\begin{table*}[ht]
    \centering
    \tablestyle{3pt}{1.2}
    \scriptsize
    \caption{\textbf{Evaluation of text-to-image generation ability on GenEval benchmark.} 
        `Gen. Only' stands for an image generation model, and `Unified' denotes a model that has both understanding and generation capabilities. 
        $\dagger$ refers to the methods using the LLM rewriter. Our model's results and BAGEL all use the LLM rewriter.}
    \begin{tabular}{lccccccc}
        \toprule
        \textbf{Model}  & \textbf{Single Obj.} & \textbf{Two Obj.} & \textbf{Counting} & \textbf{Colors} & \textbf{Position} & \textbf{Color Attri.} & \textbf{Overall$\uparrow$} \\
        \midrule
        \rowcolor{gray!10}
        \multicolumn{8}{l}{\textit{Gen. Only Models}} \\
        PixArt-$\alpha$~\cite{chen2024pixart} & 98.0 & 50.0 & 44.0 & 80.0 & 8.0 & 7.0 & 48.0 \\
        Emu3-Gen~\cite{emu3}  & 98.0 & 71.0 & 34.0 & 81.0 & 17.0 & 21.0 & 54.0 \\
        SDXL~\cite{sdxl} & 98.0 & 74.0 & 39.0 & 85.0 & 15.0 & 23.0 & 55.0 \\
        DALL-E $3$~\cite{dalle3} & 96.0 & 87.0 & 47.0 & 83.0 & 43.0 & 45.0 & 67.0 \\
        SD3-Medium~\cite{sd3} & 99.0 & 94.0 & 72.0 & 89.0 & 33.0 & 60.0 & 74.0 \\
        FLUX.1-dev$^{\dagger}$~\cite{flux} & 98.0 & 93.0 & 75.0 & 93.0 & 68.0 & 65.0 & 82.0 \\
        \midrule
        \rowcolor{gray10}
        \multicolumn{8}{l}{\textit{Unified Models}} \\
        SEED-X~\cite{seedx} & 97.0 & 58.0 & 26.0 & 80.0 & 19.0 & 14.0 & 49.0 \\
        TokenFlow-XL~\cite{tokenflow} & 95.0 & 60.0 & 41.0 & 81.0 & 16.0 & 24.0 & 55.0 \\
        ILLUME~\cite{illume} & 99.0 & 86.0 & 45.0 & 71.0 & 39.0 & 28.0 & 61.0 \\
        Transfusion~\cite{transfusion} & - & - & - & - & - & - & 63.0 \\
        Emu$3$-Gen$^{\dagger}$\cite{emu3} & 99.0 & 81.0 & 42.0 & 80.0 & 49.0 & 45.0 & 66.0 \\
        Show-o~\cite{showo} & 98.0 & 80.0 & 66.0 & 84.0 & 31.0 & 50.0 & 68.0 \\
        Janus-Pro-7B~\cite{januspro} & 99.0 & 89.0 & 59.0 & 90.0 & 79.0 & 66.0 & 80.0 \\
        MetaQuery-XL$^{\dagger}$~\cite{metaquery} & - & - & - & - & - & - & 80.0 \\
        \midrule
        \rowcolor{gray10}
        \multicolumn{8}{l}{\textit{Inference on 512 resolution}} \\
        BAGEL & 99.1 & 95.0 & 74.1 & 90.4 & 71.0 & 74.8 & 84.0 \\
        AlphaGRPO$^{\dagger}$ ({RT2I}) & 98.1 & 95.7 & 82.5 & 91.0 & 74.3 & 68.8 & 85.1 \\ 
        AlphaGRPO$^{\dagger}$ & 98.8 & 97.0 & 75.6 & 91.0 & 69.8 & 73.3 & 84.2 \\ 
        \rowcolor{myblue} 
         \textbf{\method}$^{\dagger}$ 
        & 98.4 & 96.0 & 83.1  &  93.4  & 79.8 & 79.0 & 88.3  \\
         \rowcolor{myblue} 
        \textbf{Self-\method}$^{\dagger}$ 
        & 98.4  &  97.7  &  84.4  &  93.6  &  81.5  &  81.5  &  89.5  \\
        \midrule
        \rowcolor{gray10}  
        \multicolumn{8}{l}{\textit{Inference on 1024 resolution}} \\
        BAGEL & 98.4 & 94.7 & 81.3 & 94.7 & 74.0 & 76.3 & 86.6 \\
        AlphaGRPO ({RT2I}) & 98.8 & 95.2 & 82.8 & 93.6 & 76.3 & 77.8 & 87.4 \\ 
        AlphaGRPO & 99.1 & 96.0 & 80.3 & 94.7 & 71.0 & 75.5 & 86.1 \\
        \rowcolor{myblue} 
        \textbf{\method}$^{\dagger}$  
        & 98.1 & 96.2 &  86.2 & 94.2 & 78.5 & 76.8 & 88.3 \\
        \rowcolor{myblue} 
        \textbf{Self-\method}$^{\dagger}$ 
        & 98.4 & 95.2 & 88.8 & 95.7 & 82.3 & 78.3 & 89.8 \\
    \bottomrule
    \end{tabular}
\label{tab:geneval}
\end{table*}

\paragraph{TIIF-Bench.}~\autoref{tab:tiifbench} gives the full TIIF-Bench results on short and long prompts. 
TIIF-Bench separates basic following, advanced following, and designer-oriented prompts, 
making it a useful stress test for instruction fidelity beyond object-level matching.
Both \method and Self-\method substantially improve the BAGEL baseline and AlphaGRPO on the overall short and long scores. 
The improvements are observed in basic attribute and relation following, advanced combinations
of attribute/relation/reasoning, and the designer-oriented real-world split.
The text subcategory remains challenging in absolute terms, 
but both methods increase the score by a large margin over BAGEL, 
suggesting that image-conditioned prompt likelihood provides useful supervision even for
fine-grained instruction details.
\newcommand{\thinrule}{\textcolor{black!60}{\vrule width 0.03pt}}
\newcolumntype{V}{@{\hspace{4pt}\thinrule\hspace{4pt}}}
\renewcommand{\arraystretch}{1.7} 
\setlength{\tabcolsep}{3pt}

\begin{table*}
\caption{\textbf{Performance of proprietary models and state-of-the-art open-source models on TIIF-Bench \textbf{testmini} subset.} Evaluated systems are grouped into (i) {diffusion-based} open-source models, (ii){autoregressive} open-source models, and (iii) {proprietary} models. 
The results of \method~and BAGEL are evaluated by GPT-4.1. ``Inf. SRR'' indicates executing the inference-time self-reflective refinement.}
\small
\centering
\resizebox{0.99\linewidth}{!}{
\begin{tabular}{
  >{\raggedright\arraybackslash}m{3cm}V
  *{2}{>{\centering\arraybackslash}m{0.7cm}V}
  *{8}{>{\centering\arraybackslash}m{0.65cm}V}
  *{12}{>{\centering\arraybackslash}m{0.65cm}V}
  >{\centering\arraybackslash}m{0.65cm}V
  >{\centering\arraybackslash}m{0.65cm}
}
\toprule
\multirow{3}{*}{{Model}}
  & \multicolumn{2}{cV}{\multirow{2}{*}{{Overall}}}
  & \multicolumn{8}{cV}{{Basic Following}}
  & \multicolumn{12}{cV}{{Advanced Following}}
  & \multicolumn{2}{c}{{Designer}} \\
\cmidrule(lr{.3em}){4-11} \cmidrule(lr{.3em}){12-23} \cmidrule(l){24-25}
& & &
  \multicolumn{2}{cV}{Avg}
  & \multicolumn{2}{cV}{Attribute}
  & \multicolumn{2}{cV}{Relation}
  & \multicolumn{2}{cV}{Reasoning}
  & \multicolumn{2}{cV}{Avg}
  & \multicolumn{2}{cV}{\makecell{Attribute\\+Relation}}
  & \multicolumn{2}{cV}{\makecell{Attribute\\+Reasoning}}
  & \multicolumn{2}{cV}{\makecell{Relation\\+Reasoning}}
  & \multicolumn{2}{cV}{Style}
  & \multicolumn{2}{cV}{Text}
  & \multicolumn{2}{c}{\makecell{Real\\World}} \\
\cmidrule(lr{.3em}){4-11} \cmidrule(lr{.3em}){12-23} \cmidrule(l){24-25}
& short & long &
  short & long &
  short & long &
  short & long &
  short & long &
  short & long &
  short & long &
  short & long &
  short & long &
  short & long &
  short & long &
  short & long
\\
\midrule
\addlinespace[-1pt]
\rowcolor{gray!10}
\multicolumn{25}{l}{{{Diffusion based} Open-Source Models}} \\[-1pt]
\midrule
FLUX.1 dev  &{{71.09}} &{71.78} &{83.12} &78.65& 87.05 &83.17 &{87.25} &80.39 &{75.01} &72.39 &65.79 &{68.54} & 67.07 &{73.69} &{73.84} &73.34 &{69.09} &{71.59} & 66.67 & 66.67 &43.83 &{52.83} &70.72 &{71.47} \\
SD XL           &54.96 &42.13 &65.72 &53.28 &59.33 &50.83 &77.57 &62.57 &60.32 &46.57 &49.73 &36.22 &47.82 &35.57 &56.22 &45.34 &52.59 &36.09 &73.33 &60.00 &16.83 & 0.83 &50.92 &41.59 \\
SD 3            &67.46 &66.09 &78.32 &77.75 &83.33 &79.83 &82.07 &78.82 &71.07 &74.07 &61.46 &59.56 &61.07 &64.07 &68.84 &70.34 &50.96 &57.84 &66.67 &76.67 &59.83 &20.83 &63.23 &67.34 \\
SD 3.5 L        &{71.15} &66.96 &78.34 &79.56 &79.50 &76.50 &80.96 &83.21 &72.46 &{78.71} &{67.67} &61.18 &66.46 &61.89 &73.53 &{74.15} &60.03 &61.53 &73.33 &63.33 &{70.52} &42.52 &64.43 &66.39 \\
\midrule
\addlinespace[-1pt]
\rowcolor{gray!10}
\multicolumn{25}{l}{{{AR based} Open-Source Models}} \\[-1pt]
\midrule
Llamagen &41.67 &38.22 &53.00 &50.00 &48.33 &42.33 &59.57 &60.32 &51.07 &47.32 &35.89 &32.61 &38.82 &31.57 &40.84 &47.22 &49.59 &46.22 &46.67 &33.33 &0.00 &0.00 &39.73 &35.62 \\
Show-o &59.72 &58.86 &73.08 &75.83 &74.83 &79.83 &78.82 &78.32 &65.57 &69.32 &53.67 &50.38 &60.95 &56.82 &68.59 &68.96 &66.46 &56.22 &63.33 &66.67 &3.83 &2.83 &55.02 &50.92 \\
Infinity &62.07 &62.32 &73.08 &75.41 &74.33 &76.83 &72.82 &77.57 &72.07 &71.82 &56.64 &54.98 &60.44 &55.57 &{74.22} &64.71 &60.22 &59.71 &{80.00} &{73.33} &10.83 &23.83 &54.28 &56.89 \\
JanusPro &66.50 &65.02 &79.33 &78.25 &79.33 &82.33 &78.32 &73.32 &{80.32} &79.07 &59.71 &58.82 &66.07 &56.20 &70.46 &{70.84} &67.22 &59.97 &60.00 &70.00 &{28.83} &{33.83} &65.84 &60.25 \\
\midrule
\rowcolor{gray!10}
\multicolumn{25}{l}{\textit{Inference on 512 resolution}} \\
\midrule
BAGEL & 75.21 & 78.56 & 81.73 & 86.11 & 85.50 & 88.00 & 84.99 & 85.39 & 74.69 & 84.94 & 73.66 & 77.61 & 77.68 & 81.55 & 67.77 & 76.48 & 78.58 & 77.86 & 90.00 & 90.00 & 33.03 & 40.72 & 84.70 & 82.09 \\
AlphaGRPO {(RT2I)} & 78.92 & 79.48 & 85.46 & 84.15 & 88.50 & 85.50 & 88.12 & 86.56 & 79.77 & 80.38 & 77.41 & 78.85 & 81.05 & 82.77 & 74.38 & 80.52 & 79.30 & 75.83 & 90.00 & 83.33 & 44.80 & 53.85 & 84.33 & 86.57 \\
AlphaGRPO & 79.05 & 79.50 & 85.56 & 83.32 & 89.50 & 85.50 & 85.34 & 83.60 & 81.86 & 80.86 & 77.12 & 79.87 & 78.59 & 83.95 & 71.81 & 78.94 & 83.02 & 79.36 & 86.67 & 93.33 & 51.13 & 45.25 & 83.58 & 84.70 \\
\rowcolor{myblue} 
\textbf{\method} & 85.23 & 84.81 & 87.67 & 87.44 & 90.00 & 90.50 & 88.12 & 89.45 & 84.90 & 82.38 & 84.85 & 85.39 & 84.34 & 87.52 & 84.99 & 84.95 & 86.41 & 86.02 & 93.33 & 93.33 & 66.97 & 58.82 & 88.06 & 90.30 \\
\rowcolor{myblue}
\textbf{Self-\method} & 85.13 & 84.33 & 89.69 & 87.58 & 88.00 & 92.00 & 93.56 & 87.37 & 87.50 & 83.38 & 83.73 & 85.70 & 85.17 & 86.63 & 83.05 & 90.72 & 84.65 & 82.38 & 93.33 & 93.33 & 60.63 & 57.01 & 90.30 & 86.19 \\
\midrule
\rowcolor{gray!10}
\multicolumn{25}{l}{\textit{Inference on 1024 resolution}} \\
\midrule
BAGEL & 76.42 & 76.15 & 83.44 & 83.72 & 87.50 & 89.00 & 86.03 & 84.35 & 76.77 & 77.81 & 75.16 & 76.68 & 79.34 & 83.12 & 70.38 & 74.58 & 78.36 & 75.58 & 93.33 & 86.67 & 36.20 & 40.72 & 79.85 & 73.51 \\
AlphaGRPO {(RT2I)} & 78.70 & 79.48 & 84.83 & 85.92 & 89.00 & 89.50 & 88.81 & 88.36 & 76.69 & 79.89 & 78.42 & 79.21 & 79.38 & 84.09 & 77.48 & 77.37 & 81.05 & 78.84 & 90.00 & 90.00 & 45.70 & 47.06 & 80.22 & 80.22 \\
AlphaGRPO & 77.74 & 78.09 & 85.39 & 82.90 & 89.00 & 89.50 & 87.31 & 82.96 & 79.85 & 76.25 & 75.62 & 77.49 & 79.46 & 83.32 & 73.28 & 76.13 & 76.48 & 75.56 & 90.00 & 90.00 & 42.53 & 44.80 & 81.72 & 84.33 \\
\rowcolor{myblue}
\textbf{\method} & 83.31 & 81.71 & 88.44 & 87.26 & 91.50 & 92.00 & 88.81 & 88.81 & 85.02 & 80.98 & 83.18 & 81.83 & 87.89 & 85.41 & 81.67 & 82.23 & 81.97 & 80.23 & 86.67 & 90.00 & 63.80 & 54.75 & 82.46 & 80.97 \\
\rowcolor{myblue}  
\textbf{Self-\method} & 82.66 & 82.54 & 87.38 & 87.60 & 90.50 & 93.00 & 89.80 & 85.68 & 81.86 & 84.10 & 81.49 & 83.15 & 84.33 & 83.39 & 80.88 & 84.25 & 81.16 & 84.71 & 86.67 & 83.33 & 61.09 & 59.73 & 87.69 & 84.70 \\
\bottomrule
\end{tabular}
}
\label{tab:tiifbench}
\end{table*}

\paragraph{DPG-Bench.}~\autoref{tab:dpgbench} further evaluates prompt following with the decomposed categories of DPG-Bench. 
Since BAGEL is already a strong baseline on this benchmark, 
the absolute margins are smaller than those on TIIF-Bench. 
Nevertheless, \method obtains the best 512-resolution overall score and improves the global, relation, and other categories, 
while Self-\method gives the strongest attribute score. 
At 1024 resolution, Self-\method achieves the best overall score among the compared BAGEL-based
methods. These results indicate that the reward learned from prompt
likelihood helps not only aggregate instruction following, but also the
relation- and attribute-sensitive subskills measured by DPG-Bench.
\begin{table*}[t]
    \centering
    \caption{\textbf{Evaluation of text-to-image generation ability on the DPG-Bench~\cite{dpg} benchmark.} * indicates our reproduced results.
    }
    \scriptsize
    \resizebox{0.8\linewidth}{!}{
        \begin{tabular}{lcccccc}
            \toprule
            Method & Global$\uparrow$ & Entity$\uparrow$ & Attribute$\uparrow$ & {Relation$\uparrow$} & {Other$\uparrow$} & {Overall$\uparrow$} \\
            \midrule
            \rowcolor{gray!10}
            \multicolumn{7}{l}{\textit{Gen. Only Models}} \\
            Hunyuan-DiT~\cite{hunyuandit} & 84.59 & 80.59 & 88.01 & 74.36 & 86.41 & 78.87 \\
            PixArt-$\Sigma$~\cite{chen2024pixart} & 86.89 & 82.89 & 88.94 & 86.59 & 87.68 & 80.54 \\
            DALLE3~\cite{dalle3} & 90.97 & 89.61 & 88.39 & 90.58 & 89.83 & 83.50 \\
            SD3-medium~\cite{sd3} & 87.90 & 91.01 & 88.83 & 80.70 & 88.68 & 84.08 \\
            FLUX.1-dev~\cite{flux} & 82.10 & 89.50 & 88.70 & 91.10 & 89.40 & 84.00 \\ 
            OmniGen~\cite{omnigen} & 87.90 & 88.97 & 88.47 & 87.95 & 83.56 & 81.16 \\
            \midrule
            \rowcolor{gray!10}
            \multicolumn{7}{l}{\textit{Unified Multimodal Models}} \\
            Show-o~\cite{showo} & 79.33 & 75.44 & 78.02 & 84.45 & 60.80 & 67.27 \\
            EMU3~\cite{emu3} & 85.21 & 86.68 & 86.84 & 90.22 & 83.15 & 80.60 \\
            TokenFlow-XL~\cite{tokenflow} & 78.72 & 79.22 & 81.29 & 85.22 & 71.20 & 73.38 \\ 
            Janus~\cite{janus} & 82.33 & 87.38 & 87.70 & 85.46 & 86.41 & 79.68 \\
            Janus Pro~\cite{januspro} & 86.90 & 88.90 & 89.40 & 89.32 & 89.48 & \underline{84.19} \\
            BLIP3-o 4B~\cite{blip3o} & - & - & - & - & - & 79.36 \\
            BLIP3-o 8B~\cite{blip3o} & - & - & - & - & - & 81.60 \\
            BAGEL~\cite{bagel} & 88.94 & 90.37 & 91.29 & 90.82 & 88.67 & {85.07} \\
            UniWorld-V1~\cite{uniworld} & 83.64 & 88.39 & 88.44 & 89.27 & 87.22 & 81.38 \\
            OmniGen2~\cite{omnigen2} & 88.81 & 88.83 & 90.18 & 89.37 & 90.27 & 83.57 \\
            \midrule
            \rowcolor{gray!10}
            \multicolumn{7}{l}{\textit{Inference on 512 resolution}} \\
            \midrule
            BAGEL* & 88.94 & 90.37 & 91.29 & 90.82 & 88.67 & 85.07 \\
            AlphaGRPO (RT2I) &  89.99 &  \textbf{92.20} &  88.49 &  90.89 &  89.12 &  85.98 \\
            AlphaGRPO & 84.99 &  90.93 &  91.22 &  92.51 &  90.11 & 86.25 \\
            \rowcolor{myblue} \textbf{\method}      & \textbf{93.80} & 92.17 & 90.78 & \textbf{92.95} & \textbf{93.08} & \textbf{88.08} \\
            \rowcolor{myblue} Self-\method & 92.40 & 91.94 & \textbf{92.12} & 92.77 & 88.57 & 87.73 \\
     
            \midrule  
            \rowcolor{gray!10}
            \multicolumn{7}{l}{\textit{Inference on 1024 resolution}} \\
            \midrule
            BAGEL* & 87.42 & \textbf{92.46} & 90.75 & 91.92 & 84.96 & 85.17 \\
            AlphaGRPO (RT2I) & 87.42 & \textbf{92.46} & 90.75 & 91.92 & 84.96 & 85.87 \\
            AlphaGRPO & \textbf{89.21} & 89.43 & 90.20 & \textbf{92.26} & 90.39 & 85.08 \\
            \rowcolor{myblue} \textbf{\method}            & 85.54 & 91.95 & 92.67 & 91.27 & 89.14 & 86.46 \\
            \rowcolor{myblue} \textbf{Self-\method} & 85.33 &  91.23 & \textbf{93.14} &  89.97 & \textbf{92.06} & \textbf{86.97}  \\
            \bottomrule
        
        \end{tabular}
    }
    \vspace{-4pt}
\label{tab:dpgbench}
\end{table*}

\paragraph{WISE.}~\autoref{tab:wisescore} reports the WISE benchmark, which emphasizes
world-knowledge and semantic reasoning in text-to-image generation. 
When combined with self-CoT, however, both methods clearly improve over BAGEL and AlphaGRPO with the same inference strategy. 
Self-\method with self-CoT reaches the best overall WISE score and improves multiple knowledge domains, 
including space, biology, physics, and chemistry. 
This suggests that \method improves prompt-following ability and helps the model respond better to the CoT text.
\begin{table*}[!ht]
    \centering
    \scriptsize
    \setlength{\tabcolsep}{8pt}
    \caption{\textbf{Comparison of world knowledge reasoning on WISE.} WISE examines the complex semantic understanding and world knowledge for T2I generation. `Gen. Only' stands for an image generation model, and `Unified' denotes a model that has both understanding and generation capabilities.
    }
    \resizebox{\textwidth}{!}{%
    \begin{tabular}{clccccccc}
    \toprule
    \textbf{Type} & \textbf{Model} & \textbf{Cultural}  & \textbf{Time}     & \textbf{Space}    & \textbf{Biology}    & \textbf{Physics} & \textbf{Chemistry} & \textbf{Overall$\uparrow$} \\
    \midrule
    \multirow{4}{*}{\rotatebox{90}{\textit{Gen. Only}}}
    & SDXL~\cite{sdxl} &  0.43  & 0.48 &0.47  &0.44  &0.45 &0.27 & 0.43 \\
    & SD3.5-large~\cite{sd3} & 0.44 &0.50 &0.58  & 0.44&0.52 &0.31 & 0.46 \\
    & PixArt-Alpha~\cite{chen2024pixart} & 0.45  & 0.50& 0.48 & 0.49& 0.56 &0.34 & 0.47\\
    & FLUX.1-dev~\cite{flux} & 0.48  &0.58 &0.62  &0.42  &0.51 & 0.35 & 0.50 \\
    \midrule
    \multirow{10}{*}{\rotatebox{90}{\textit{Unified}}} 
    & Janus~\cite{janus} &0.16 &0.26 &0.35 & 0.28 &0.30 & 0.14& 0.23\\
    & VILA-U~\cite{vilau} & 0.26 &0.33  & 0.37 &0.35  &0.39 &0.23 & 0.31\\
    & Show-o-512~\cite{showo} & 0.28 &0.40  &0.48 & 0.30& 0.46 & 0.30 & 0.35\\
    & Janus-Pro-7B~\cite{januspro} & 0.30& 0.37& 0.49 & 0.36&0.42 &0.26 & 0.35 \\
    & Emu3~\cite{emu3} & 0.34 & 0.45 & 0.48 & 0.41  & 0.45 & 0.27 & 0.39 \\
    & MetaQuery-XL~\cite{metaquery} & 0.56& 0.55 &0.62 &  0.49 &  0.63 & 0.41 & 0.55 \\
    & {BAGEL}~\cite{bagel} & 0.44 & 0.55 & 0.68 & 0.44 & 0.60 & 0.39 & 0.52 \\
    & {BAGEL} w/ Self-CoT~\cite{bagel} & 0.76 & 0.69 & 0.75 & 0.65 & 0.75 & 0.58 & {0.70} \\
    & AlphaGRPO & 0.44 & 0.55  & 0.64  & 0.46  & 0.62  & 0.46 & 0.53 \\
    & AlphaGRPO~w/ Self-CoT & 0.75 & 0.70 & 0.74 & 0.66 & 0.77 & 0.64 & 0.71 \\
    \rowcolor{myblue} & \textbf{\method} & 0.42 & 0.51 & 0.67 & 0.48 & 0.65 & 0.47 & 0.53 \\
    \rowcolor{myblue} & \textbf{\method}~w/ Self-CoT & 0.76 & 0.70 & 0.80 & 0.72 & 0.80 & 0.65 & 0.74 \\
    \rowcolor{myblue} & \textbf{Self-\method} & 0.40 & 0.53 & 0.67 & 0.43 & 0.64 & 0.45 & 0.52 \\
    \rowcolor{myblue} & \textbf{Self-\method}~w/ Self-CoT & 0.75 & 0.71 & 0.82 & 0.73 & 0.84 & 0.69 & 0.76 \\
    \bottomrule
    \end{tabular}
    }
    \label{tab:wisescore}
\end{table*}

\section{More Visualizations}
\label{sec:more_visualizations}

\begin{figure*}[t]
    \centering
    \includegraphics[width=\linewidth]{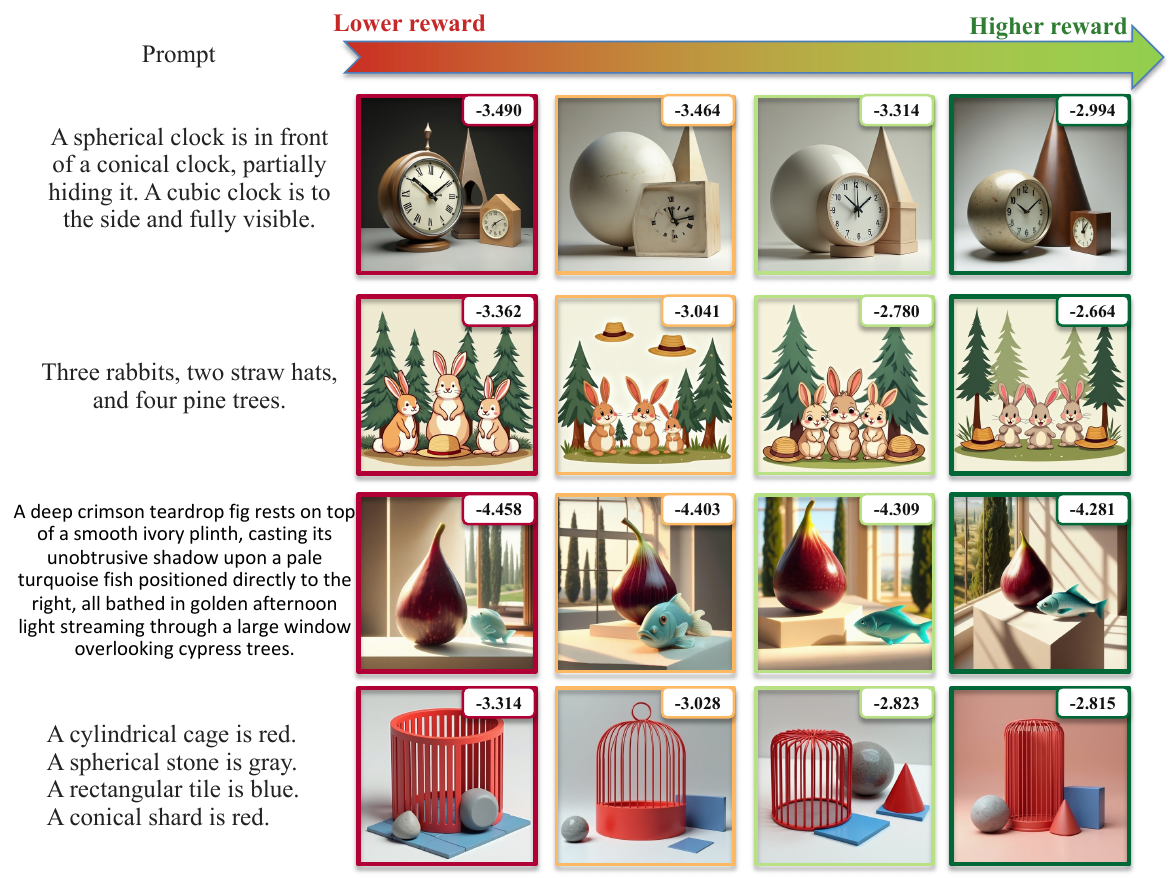}
    \caption{\textbf{Additional reward-ranking examples.}
            For each prompt, samples are ordered from lower to higher \method reward.}
    \label{fig:appendix_reward_ranking}
\end{figure*}

\autoref{fig:appendix_reward_ranking} provides additional examples of reward ranking in the same rollout group. 
Samples with higher \method reward better satisfy object, attribute, and spatial constraints, 
supporting the use of image-conditioned prompt likelihood as a group-wise ranking signal.

\begin{figure*}[t]
    \centering
    \includegraphics[width=\linewidth]{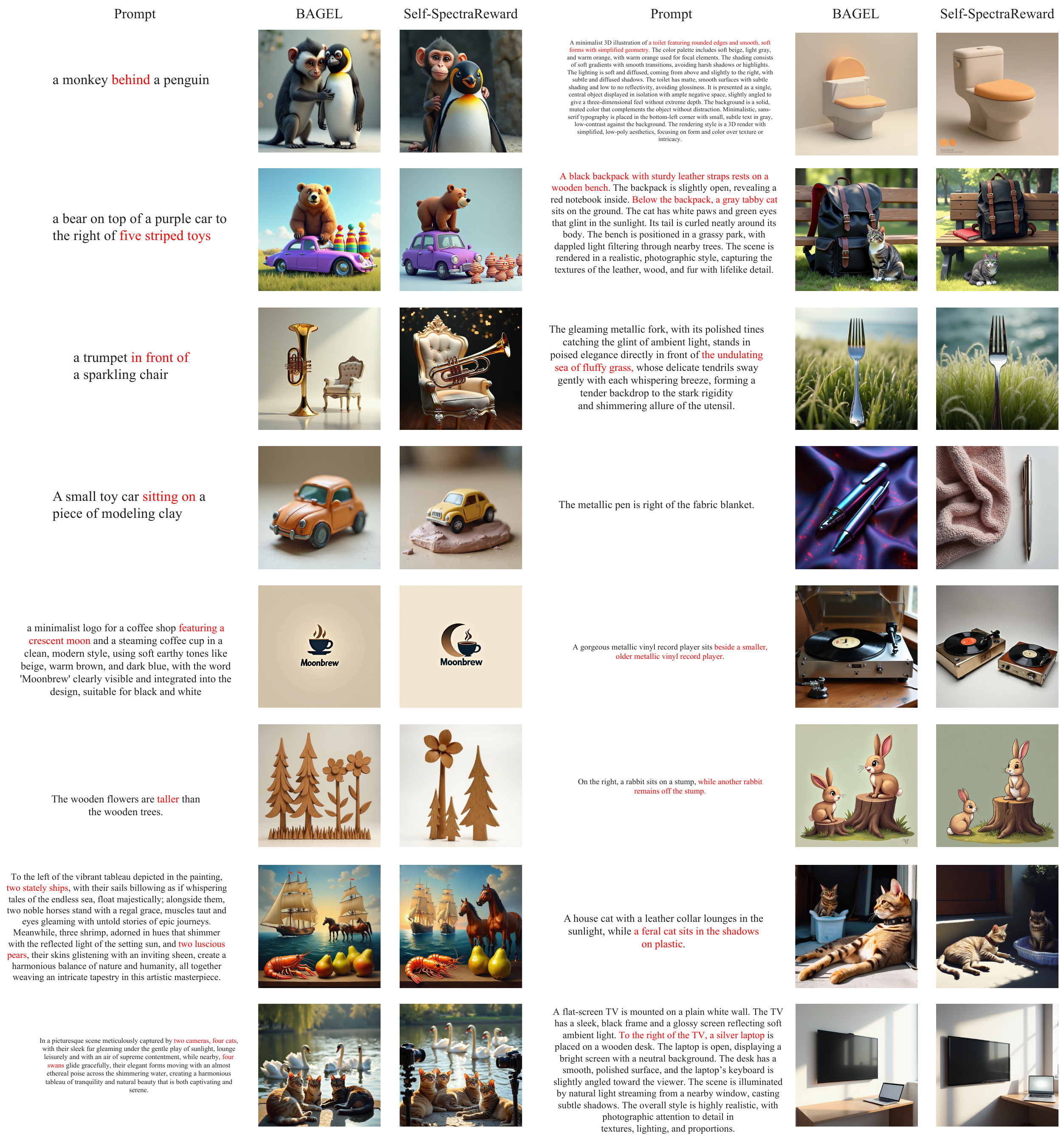}
    \caption{\textbf{Additional qualitative comparison.}
    Red text highlights the prompt constraints that are especially sensitive to spatial relation,
    counting, attribute binding, or long-prompt composition.
    Self-\method more consistently satisfies these highlighted constraints than the BAGEL baseline.}
    \label{fig:appendix_qualitative}
\end{figure*}

\autoref{fig:appendix_qualitative} provides additional qualitative
comparisons between the BAGEL baseline and Self-\method. The red text marks the key differences between the baseline BAGEL and our model, 
covering spatial relations, counting, attribute binding, relative size, and object co-occurrence in long prompts.
Across these cases, Self-\method more often preserves the requested relation between objects while also keeping the main visual content recognizable. 
For example, it places the toy car on modeling clay and 
separates the pen from the fabric blanket. These examples support the quantitative gains in
instruction-following benchmarks by showing that the learned reward improves fine-grained prompt grounding rather than only global image quality.

\section{Broader Impacts}\label{sec:broader_impacts}
\method lowers the cost of constructing reward signals for image-generation RL by reusing frozen pretrained MLLMs, 
without preference annotation or reward-model fine-tuning. 
This may make reinforcement learning for text-to-image generation more accessible to researchers with limited annotation or
training resources, and can support studies on prompt following, compositional generation, and unified multimodal self-improvement.

At the same time, easier RL optimization for image generation may amplify the existing risks of text-to-image models. 
Stronger prompt following can be misused for generating misleading visual content, imitating protected styles, 
or producing unsafe images. Since \method derives rewards from pretrained MLLMs, 
it may inherit potential biases and visual misjudgments from the reward MLLM backbone during policy optimization. 
For Self-\method, the closed-loop reward further encourages the model to align with its own understanding branch,
which may not always reflect human preferences or safety requirements. 
We therefore recommend using \method together with standard safety filters, bias auditing, and human evaluation 
when deploying or releasing models trained with this reward.


\end{document}